\documentclass[journal]{IEEEtran}
\usepackage{stfloats}
\usepackage{comment}
\usepackage{enumitem}  

\usepackage[utf8]{inputenc}  
\usepackage[table]{xcolor} 

\definecolor{verylightgray}{rgb}{0.92,0.92,0.92}
\usepackage{authblk}
\usepackage{graphicx}
\usepackage{xcolor}
\usepackage{cite}
\usepackage{booktabs}
\usepackage[pagewise]{lineno}
\usepackage{balance}
\usepackage{amssymb}
\usepackage{url}
\usepackage{booktabs}
\usepackage{graphicx}
\usepackage{array}
\usepackage{multirow}
\usepackage{amsmath}
\usepackage[table]{xcolor}
\usepackage{array}
\usepackage{longtable}
\usepackage{supertabular} 
\usepackage{threeparttable}

\usepackage{pifont}
\usepackage[table,xcdraw]{xcolor}

\title{LiDAR for Rehabilitation: A Comprehensive Survey of Applications, AI Techniques, and Future Directions}

\author{
\IEEEauthorblockN{Soumia Siyoucef, \textit{Student Member, IEEE}, Najmeddine Dhieb, \textit{Member, IEEE}, Hakim Ghazzai, \textit{Senior Member, IEEE}, 	Eleonora Guanziroli, Franco Molteni, and Gianluca Setti, \textit{Fellow, IEEE}}\\
{\thanks {\hrule
\vspace{0.1cm} 
Soumia Siyoucef, Najmeddine Dhieb, Hakim Ghazzai, and Gianluca Setti are with the Computer, Electrical and Mathematical Sciences \& Engineering (CEMSE) Division at King Abdullah University of Science and Technology (KAUST), Thuwal, Saudi Arabia (E\textendash mails: \{soumia.youcef, najmeddine.dhieb, hakim.ghazzai, gianluca.setti\}@kaust.edu.sa).\\
Eleonora Guanziroli and Franco Molteni are with Valduce Hospital, Villa Beretta Rehabilitation Center, Costa Masnaga, Italy.

This paper is accepted for publication in  IEEE Sensors Reviews, April, 2026.
Personal use of this material is permitted.  Permission from IEEE  must be obtained for all other uses, in any current or future media, including reprinting/republishing this material for advertising or promotional purposes, creating new collective works, for resale or redistribution to servers or lists, or reuse of any copyrighted component of this work in other works.
}}}

\date{\today}

\graphicspath{{figures/}}

\begin{document}

\maketitle

\begin{abstract}
Rehabilitation aims to help patients with limited mobility regain their physical abilities through targeted movements, exercises, stimulation, and other therapeutic methods. Recent advances in technology have introduced sensor‑based systems into rehabilitation and clinical practices, enabling real‑time monitoring and providing accurate feedback on movement accuracy. Among these sensors, LiDAR has demonstrated strong potential, offering key advantages over conventional techniques such as camera‑based systems, which raise privacy concerns, and wearable sensors, which can be uncomfortable and prone to errors. In this work, we review the applications of LiDAR in rehabilitation, post‑injury care, and hospital environments, focusing on studies published between 2019 and 2025. Studies across several areas have been explored: 3D body scanning and gait analysis with standalone LiDAR, LiDAR mounted on robotic systems for rehabilitation, real‑time monitoring and environment scanning for safe navigation, and activity and position recognition. We also analyze processing techniques, particularly learning‑based approaches, and support the discussion with statistical analysis,  highlighting trends, gaps, and future research opportunities. To the best of our knowledge, this is the first comprehensive survey dedicated to LiDAR for rehabilitation applications, providing an overview of current methods, AI‑based processing techniques, and open challenges.

\end{abstract}


\section{Introduction}

\textcolor{black}{Rehabilitation is defined by the World Health Organization (WHO) as a set of interventions that help patients regain their ability to carry out specific daily tasks following injuries, neurological conditions such as Parkinson’s disease or stroke, postoperative recovery, or the natural effects of aging, such as walking, grasping objects, or moving joints smoothly~\cite{mills2017rehabilitation}. Conventional rehabilitation techniques include manual therapy and exercise-guided therapy. Manual therapy involves caregivers performing movements directly on the patient, such as muscle massage, stretching, and joint mobilization~\cite{cho2020effects}. This approach is typically used in severe cases, where the patient’s mobility is too limited to perform movements independently, or during the early stages of rehabilitation. As the patient progresses, exercise-guided rehabilitation can be introduced~\cite{salvi2018m}, in which patients perform movements themselves under supervision. At this stage, supportive devices such as parallel bars, walkers, canes, and grab bars can assist the process~\cite{brummel2009rehabilitation}. Rehabilitation can also be delivered remotely through tele-rehabilitation~\cite{salvi2018m}. Other approaches include functional electrical stimulation to restore neural function in impaired muscles using low-power electrical signals~\cite{maffiuletti2018clinical}, heat therapy~\cite{celi2024effectiveness}, and mechanical stimulation based on haptic feedback, such as pressure, vibration, taping, and resistance~\cite{ferre2011haptic}.} More advanced approaches may involve neurological stimulation and virtual/augmented reality. The integration of technology into medical devices has significantly advanced the field of rehabilitation in several ways~\cite{boltaboyeva2025review}. In fact, traditional physiotherapy, which requires physical contact between caregivers and patients, can now be complemented by specialized devices and robots equipped with sensors to assess exercise progress and actuators that apply controlled forces to assist patients in performing specific movements targeting impaired limbs~\cite{8572762}. Wearable sensor systems, placed on the patient’s body, help monitor exercise progress and track goals such as joint angle measurements. Another example of advanced healthcare technology is exoskeletons~\cite{faridi2022machine,yao2024advancements,secciani2021wearable,11319164}, which serve a dual purpose: they assist in rehabilitation by supporting the patient, triggering movements, and providing neurological stimulation. Additionally, exoskeletons act as daily assistive devices that replace the function of a body part in cases where recovery and regaining mobility are not possible.
\begin{table*}[!t]
\centering
\caption{Summary of recent survey paper relevant to the use of LiDAR in rehabilitation and medical applications. The table outlines the scope of each paper and indicates whether it: (A) focuses on the use of LiDAR/point clouds, (B) targets healthcare and rehabilitation applications, (C) discusses AI integration for point cloud processing, (D) examines human body measurements such as scanning and gait assessment, (E) explores assistive robots in the medical field, and (F) addresses activity recognition.}
\resizebox{\textwidth}{!}{
\begin{tabular}{|@{\hskip3pt}p{7cm}@{\hskip3pt}|
@{\hskip3pt}p{0.7cm}@{\hskip3pt}|
@{\hskip3pt}p{7cm}@{\hskip3pt}|
@{\hskip3pt}p{0.4cm}@{\hskip3pt}|
@{\hskip3pt}p{0.4cm}@{\hskip3pt}|
@{\hskip3pt}p{0.4cm}@{\hskip3pt}|
@{\hskip3pt}p{0.4cm}@{\hskip3pt}|
@{\hskip3pt}p{0.4cm}@{\hskip3pt}|
@{\hskip3pt}p{0.4cm}@{\hskip3pt}|}
\hline 
\textbf{Title} & \textbf{Year} & \textbf{Scope} & \textbf{A} & 
\textbf{B} & \textbf{C} & \textbf{D} & \textbf{E} & \textbf{F} \\

\hline
LiDAR-based detection, tracking, and property estimation: A contemporary review~\cite{hasan2022lidar} & 2022 & Focus on LiDAR-based human pose estimation & \textcolor{green}{\ding{51}} & \textcolor{red}{\ding{55}} & \textcolor{green}{\ding{51}} &\textcolor{red}{\ding{55}}   & \textcolor{green}{\ding{51}} & \textcolor{red}{\ding{55}}   \\
\hline
A Review of Intelligent Walking Support Robots: Aiding Sit-to-Stand Transition and Walking~\cite{sun2024review} & 2024 & Review the available intelligent robots for walking and sit-to-stand aiding. focus on control strategies and safety & \textcolor{red}{\ding{55}} & \textcolor{green}{\ding{51}} & \textcolor{red}{\ding{55}} & \textcolor{red}{\ding{55}} & \textcolor{green}{\ding{51}} & \textcolor{red}{\ding{55}} \\ 
\hline
Advancements in Sensor Technologies and Control Strategies for Lower-Limb Rehabilitation Exoskeletons: A Comprehensive Review~\cite{yao2024advancements} & 2024 & Review sensors and control techniques of Lower limb exoskeletons & \textcolor{red}{\ding{55}} & \textcolor{green}{\ding{51}} & \textcolor{red}{\ding{55}} & \textcolor{red}{\ding{55}} & \textcolor{green}{\ding{51}} & \textcolor{red}{\ding{55}} \\ 
\hline
Study of Human–Robot Interactions for Assistive Robots Using Machine Learning and Sensor Fusion Technologies~\cite{raj2024study} & 2024 & Review the human-robot interaction in assistive robots, Sensors and \textcolor{black}{artificial intelligence (AI)} techniques & \textcolor{red}{\ding{55}} & \textcolor{green}{\ding{51}} & \textcolor{red}{\ding{55}} & \textcolor{red}{\ding{55}} & \textcolor{green}{\ding{51}} & \textcolor{red}{\ding{55}} \\ 
\hline
Towards an ultrafast 3D imaging scanning LiDAR system: a review~\cite{li2024towards} & 2024 & Summarized LiDAR-based 3D imaging techniques, focus on the scanning speed & \textcolor{green}{\ding{51}} & \textcolor{red}{\ding{55}} & \textcolor{red}{\ding{55}} & \textcolor{green}{\ding{51}} & \textcolor{red}{\ding{55}} & \textcolor{red}{\ding{55}} \\ 
\hline
3D ToF LiDAR for Mobile Robotics in Harsh Environments: A Review~\cite{yang20253d} & 2025 & Summarizes different uses of 3D ToF LiDARs for mobile robots, including medical assistant robots & \textcolor{green}{\ding{51}} & \textcolor{red}{\ding{55}} & \textcolor{green}{\ding{51}} & \textcolor{red}{\ding{55}} & \textcolor{green}{\ding{51}} & \textcolor{red}{\ding{55}} \\
\hline  

3D Human Pose and Shape Estimation from LiDAR Point Clouds: A Review~\cite{galaaoui20253d} & 2025 & Review LiDAR-based 3D human pose and shape estimation using point clouds and AI methods& \textcolor{green}{\ding{51}} & \textcolor{red}{\ding{55}} & \textcolor{green}{\ding{51}} & \textcolor{green}{\ding{51}} & \textcolor{red}{\ding{55}} & \textcolor{green}{\ding{51}} \\
\hline 

\textbf{This work} & 2025 & Summarizes all different applications of LiDAR in health care and rehabilitation & \textcolor{green}{\ding{51}} & \textcolor{green}{\ding{51}} & \textcolor{green}{\ding{51}} & \textcolor{green}{\ding{51}} & \textcolor{green}{\ding{51}} & \textcolor{green}{\ding{51}} \\ 
\hline

\end{tabular}}
\vspace{-0.5cm}

\label{surveys}
\end{table*}

A major limitation of conventional rehabilitation methods is the lack of proper feedback on the correctness of movements performed. Which has led researchers to explore innovative ways to monitor rehabilitation exercises and precisely assess patient's movements. Recently, the use of LiDAR for this purpose has gained increasing interest. \textcolor{black}{LiDAR, which refers to light detection and ranging, is an optical sensing technology that uses laser pulses to measure distances by calculating the time required for the emitted light to return after reflecting off objects, enabling the capture of the environment’s shape and structural features.} This novel integration allows for precise assessment of movements in three-dimensional space, maintains robust performance under various environmental conditions, and protects patient privacy by avoiding the collection of identifiable data~\cite{zhao2024lidar}.

LiDAR-generated output is visually interesting, capturing multi-frame data in point cloud format and accurately representing object shapes. It allows to remove distractions such as  colors, facial features, or background details, and focuses only on shapes and movements in dense 3D coordinate points. It can also provide information on textures by leveraging signal intensity and material reflectivity, showing detailed motion information with frame rates of up to 60–100 frames per second. This sparse, multidimensional data has been widely used as input to AI models given its large and complex nature~\cite{camuffo2022recent}. AI not only reduces it to a smaller set of features that retain essential information but also supports regression tasks such as estimating  joint angles, body measurements, movement speed, and others.
Additionally, numerous classification models are available in the literature that categorize point cloud data into body part categories, helping segment the scanned body and even determine the person’s position in real-time. In most rehabilitation processes, the detection of body key joints is essential. In conventional approaches, this task is typically performed in movement analysis laboratories, for example by placing reflective markers on joints and capturing motion data, followed by image processing techniques to extract a skeletal model and measure joint angles over time~\cite{moeslund2006survey}. \textcolor{black}{In contrast, LiDAR-based approaches make this process less manual and more robust~\cite{yoon2021development}. In addition, recent AI models are able to estimate 2D and 3D joint positions and track the human skeleton in real time, achieving high accuracy across various datasets \textcolor{black}{on healthy participants}~\cite{zhang2024neighborhood,ren2024livehps}.}

Recent interest in using LiDAR in rehabilitation focuses on two main applications. One is using LiDAR for static body part or full-body scans to monitor the progress of rehabilitation over sessions~\cite{oberhofer2024feasibility,lay2023preventing}. The other application is scanning the body during rehabilitation sessions to provide real-time feedback on whether the movements are being performed correctly. In these cases, LiDAR can either be used alone in a fixed position or mounted on a mobile robot that follows the patient’s walking or movements~\cite{lee2020development,tan2023quantitative}. LiDAR is also commonly used in rehabilitation robots for other purposes, such as scanning the surrounding environment to support navigation and patient assistance~\cite{ibrayev2024development,eun2023self,huhs2025application}. 
\textcolor{black}{The effective integration of LiDAR into rehabilitation applications requires attention to many aspects. First, the environment where patient monitoring is carried out should account for the sensitivity of LiDAR technology, avoiding reflective materials, transparent surfaces, and disturbing light sources~\cite{henley2023detection}. Additionally, the choice of the sensor setup is important, whether using LiDAR alone, multiple LiDARs, or LiDAR combined with other sensors it must be accompanied by appropriate processing steps, starting from spatial alignment of the readings and temporal synchronization, to the design of an adequate pipeline that handles the multisource data~\cite{barcelo2019self, leong2024lidar}. It additionally brings challenges in terms of storage needs, processing time, computational resources, and energy consumption.}

To the best of our knowledge, there is no dedicated survey that specifically investigates the use of LiDAR technology in rehabilitation applications. Existing review papers either focus on other domains outside the healthcare context or discuss LiDAR in conjunction with additional sensing modalities such as cameras and \textcolor{black}{inertial measurement units (IMUs)}, without isolating LiDAR’s unique role and potential. Table~\ref{surveys} provides an overview of the most relevant survey papers, outlining their scope and limitations for readers seeking insights specifically into LiDAR-based rehabilitation and healthcare solutions.

\begin{figure*}[h!]
    \centering \includegraphics[width=1\linewidth]{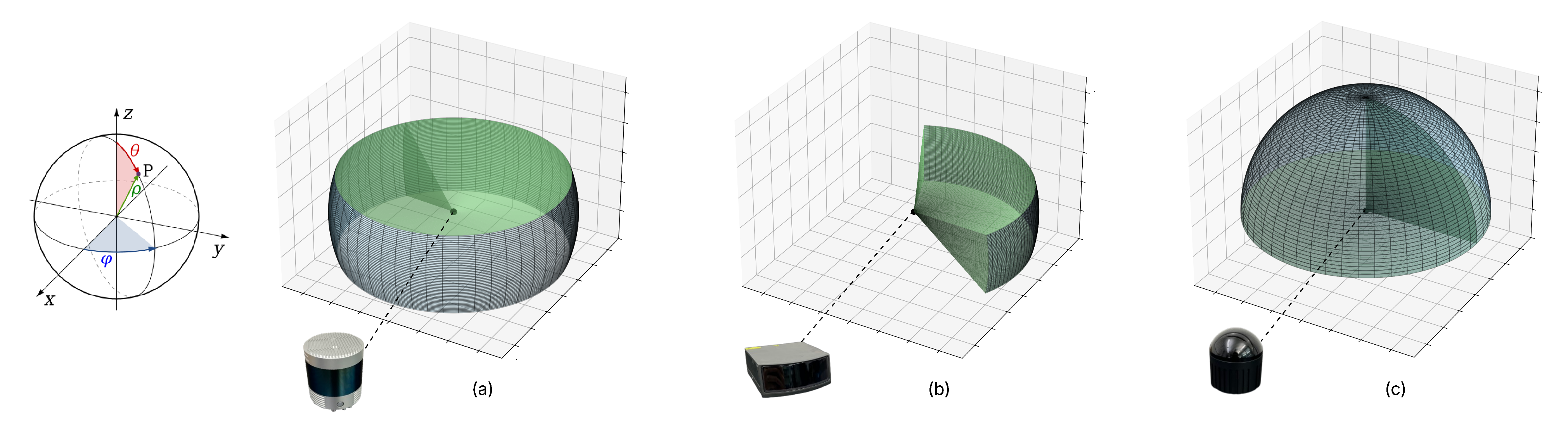}
    \caption{Different LiDAR scanning configurations:  
(a) 360° azimuth coverage with limited elevation range 
$\theta \in [\theta_{1}, \theta_{2}]$, where $0 \ll \theta_{1} < \tfrac{\pi}{2}$ and $\tfrac{\pi}{2} < \theta_{2} \ll \pi$;  
(b) Front-facing LiDAR with limited elevation and azimuth ranges,  
$\theta \in [\theta_{1}, \theta_{2}]$ with $0 \ll \theta_{1} < \tfrac{\pi}{2}$ and $\tfrac{\pi}{2} < \theta_{2} \ll \pi$,  
and $\phi \in [\phi_{1}, \phi_{2}]$ with $\phi_{2} - \phi_{1} \leq \pi$;  
(c) Half-spherical coverage with $\phi \in [0, 2\pi]$ and $\theta \in [0, \tfrac{\pi}{2}]$.}
    \label{types}
    \vspace{-0.5cm}
\end{figure*}
\textcolor{black}{We selected representative review papers related to our topic, including rehabilitation monitoring, the use of perception devices such as LiDAR and cameras, and assistive robotic systems, in order to position our work within the existing literature. We focused on the most relevant and recent reviews, identified based on keyword similarity to our topic.} Yang et al.~\cite{yang20253d}, in 2025, summarized the application of 3D LiDARs in mobile robots operating in harsh environments, including hospital settings. Li et al.\cite{li2024towards} in 2024 \textcolor{black}{conducted} a survey summarizing 3D scanning techniques using LiDAR, as do the review papers by Raj et al.\cite{raj2020survey} in 2020 and Bi et al.~\cite{bi2021survey} in 2021, each focusing on specific aspects of LiDAR-based scanning, such as speed and cost. Other available reviews addressed algorithms for processing LiDAR point clouds, such as the work by Camuffo et al.\cite{camuffo2022recent} in 2022, which highlighted recent advances in AI-based point cloud processing. Similarly, Xu et al.\cite{xu2021review} provided an overview of techniques for point cloud-based pose estimation, and Hasan et al.~\cite{hasan2022lidar} followed a similar direction in their 2022 study. While these surveys concentrated on LiDAR technology and its processing algorithms, none of them specifically target medical applications and rehabilitation scenarios.

Moreover, the reviews by Bartol et al.~\cite{bartol2021review} and Paoli et al.~\cite{paoli2020sensor} provided a broader overview of 3D scanning sensors, although they are not specifically focused on LiDAR. More recent published volumes, such as \emph{Technology for Inclusion and Participation for All}~\cite{mavrou2025technology} and the \textcolor{black}{ \emph{proceedings of the IoT and LiDAR technologies in healthcare workshop (ILTH 2024)}}~\cite{Suryadevara2025ILTH}, summarize a wide range of contemporary research and applications in rehabilitation and healthcare, including the use of LiDAR in combination with other IoT technologies. Finally, several 2024 review articles address medical support robots from complementary perspectives, including LiDAR-based control strategies and safety~\cite{sun2024review}, sensing technologies~\cite{yao2024advancements}, and human--robot interaction~\cite{raj2024study}.

Therefore, in this paper, we present the first and most comprehensive survey dedicated to the use of LiDAR technology in rehabilitation. \textcolor{black}{The target audience of this review includes researchers, clinicians, and system designers working in LiDAR and human motion analysis, as well as rehabilitation practitioners seeking to integrate sensing technologies into their practice.} Our work systematically reviews recent research efforts where LiDAR plays a central role in sensing, assessment, and therapeutic support across various rehabilitation contexts. By focusing exclusively on LiDAR-based systems, we aim to fill an important gap in the literature and provide a valuable reference for researchers, clinicians, and system designers working at the intersection of LiDAR sensing and rehabilitation technology. \textcolor{black}{Hence, in this paper, we provide a global overview of LiDAR technology, including its underlying principles and the structure of the generated data. We compare LiDAR with other sensing technologies used in rehabilitation and review the use of AI techniques for processing LiDAR data.} We then focus on its applications in rehabilitation, post-injury care, and hospital environments. In particular, we cover the following areas: the use of LiDAR as a standalone device for static body scanning or gait analysis, its integration in robotic systems, either to monitor rehabilitation in real-time or to assist patient navigation, its use for position and activity recognition, as well as other miscellaneous applications. We then summarize the key findings, highlight emerging trends, and present a statistical analysis that exposes several gaps that remain to be addressed. We also provide a detailed analysis of AI-based processing techniques, examining the common pipelines and architectures used to process LiDAR point cloud data, along with the available datasets that serve as benchmarks for evaluating these models. Finally, we highlight open research areas and future perspectives, outlining key challenges that remain to be addressed. 

\textcolor{black}{The rest of the paper is organized as follows. Section~II introduces LiDAR technology, explaining its working principle and providing an overview of the AI techniques used to process point cloud data across different stages. Section~III provides a comprehensive overview of LiDAR-based studies in the different aforementioned rehabilitation applications. A statistical analysis of current trends is then presented in Section~IV. Section~V discusses the available datasets related to the use of LiDAR sensors in rehabilitation. Afterwards, the open research directions and challenges that need to be addressed in the field are outlined in Section VI. Finally, the paper is concluded in Section~VII.}
\section{LiDAR Technology Overview and AI Techniques}
This section\footnote{This section is intended for readers who are not familiar with LiDAR technology. Those already acquainted with it may proceed directly to Section~III.} introduces LiDAR technology, explaining how it is used to sense the surrounding environment and how its data can be processed using AI techniques. It also provides a qualitative comparison between LiDAR and other sensing modalities, highlighting its advantages and limitations in the context of rehabilitation.
\subsection{LiDAR Technology}
LiDAR is an optical sensor used to map the surrounding or facing environment~\cite{lidar2022lidar}. It operates by emitting laser light waves and receiving their reflections. The reflected signals are analyzed internally to determine the distance the light has traveled~\cite{lidar2022lidar}. LiDAR can emit light waves either continuously or in discrete pulses at various angles very rapidly, allowing it to scan the environment efficiently. It can be stationary, measuring distances relative to itself, or mobile, such as when mounted on aircraft for ground mapping. \textcolor{black}{LiDAR technology can achieve relatively long ranges by relying on directive, narrow laser beams that travel over several meters and reflect back within microseconds, enabling high-precision scans.}

\textcolor{black}{
\textcolor{black}{Time-of-flight (ToF)} is the most commonly used distance measurement principle with LiDAR. It  works by sending out laser pulses and measuring the time it takes for each pulse to return after reflecting off an object~\cite{chaudhari2023fmcw}. This measured time is used to calculate the distance between the emitter and the reflection point, enabling the collection of discrete points that represent the depth of the environment in front of the LiDAR. While the ToF technique uses discrete laser pulses, continuous wave LiDAR employs a continuous laser beam and analyzes the phase and/or frequency shifts in the returning signal~\cite{chaudhari2023fmcw}, resulting in two types of LiDARs: frequency-modulated and phase-modulated. In frequency modulated continuous wave LiDARs, the returned signal is mixed with the original laser frequency. This operation allows detecting the frequency shifts, which reveal how far the object is and how fast it is moving~\cite{chaudhari2023fmcw}. Phase modulated continuous wave LiDAR uses changes in the signal’s phase to measure distance. The system sends a laser beam modulated by a repeating pulse pattern, often shaped like a square wave. When the signal reflects off the target and returns, it is compared with the original pattern~\cite{chaudhari2023fmcw}. Phase shift between the sent and received signals is related to the distance the light traveled, allowing the system to calculate how far the object is.}
\begin{figure}[h!]
    \centering \includegraphics[width=0.9\linewidth]{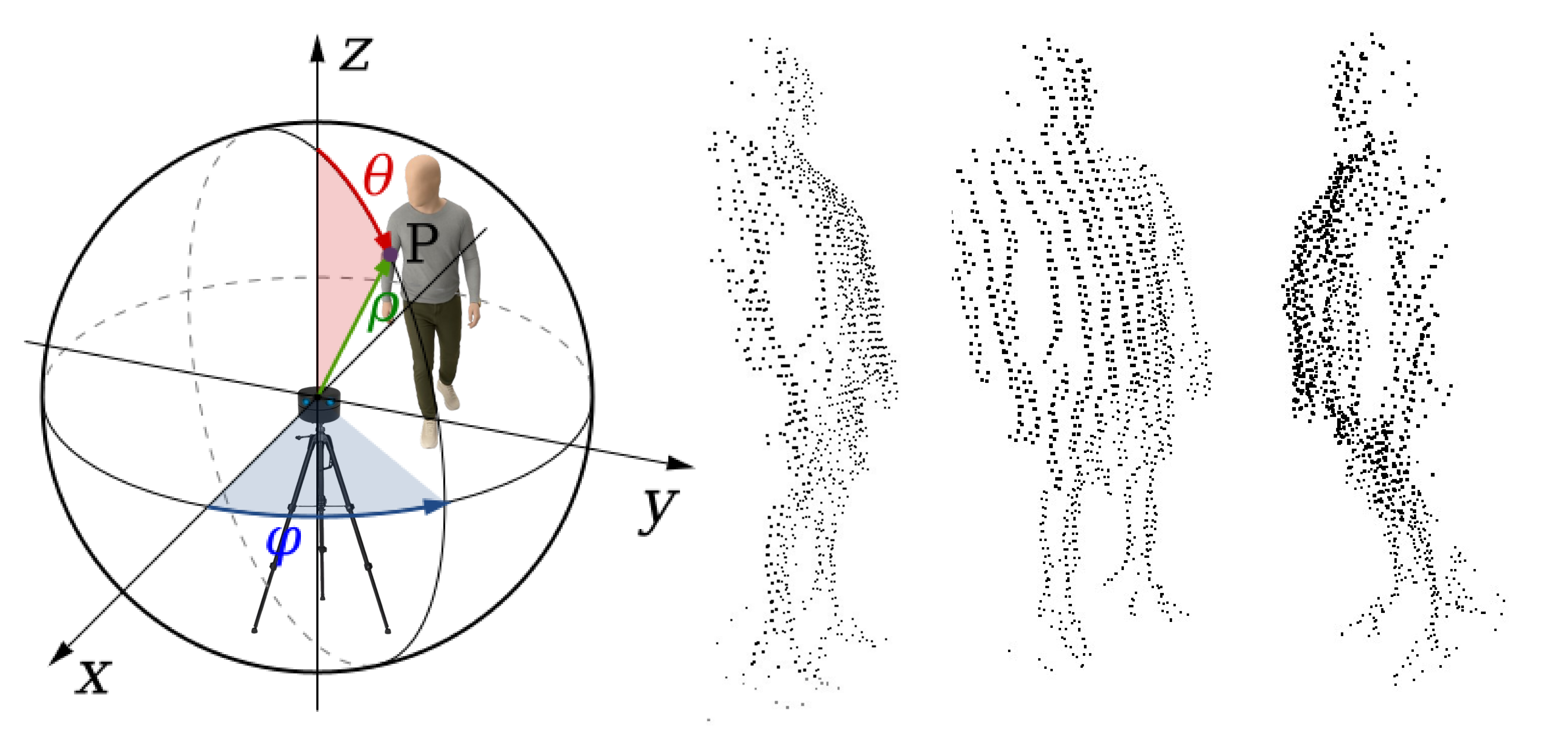}
    \caption{On the left, a single laser beam emitted by the LiDAR hits the target, generating one point in the point cloud corresponding to a specific azimuth ($\theta$) and elevation ($\phi$) angle. By emitting multiple beams spanning the LiDAR's azimuth ($\theta$) and elevation ($\phi$) ranges, a dense set of points is collected, forming the complete 3D point cloud, as illustrated at the right.
}
    \label{cartesian}
    \vspace{-0.2cm}
\end{figure}

The main challenges often reported in the use of LiDAR are its slight sensitivity to environmental settings, such as lighting conditions and the reflectivity of materials~\cite{kim2023study,akulovas2025development}. Given its reliance on reflected laser light beams, it is prone to minor noise and interference from other light sources. Second, the occlusion effect: LiDAR fails to scan any objects or parts of objects that are hidden behind obstacles, which is often uncontrollable in many scenarios~\cite{wu2024lidar,ren2024livehps,saroja2021human}. For example, when scanning legs, the right and left legs may hide each other from the side view~\cite{duong2020human}. To overcome this occlusion, many studies rely on the use of multiple LiDAR sensors from different viewpoints. This solution can mitigate occlusion but requires calibration between the two readings in the spatial domain, as well as synchronization~\cite{duong2021human,duong2024self,jiao2019novel}. Moreover, using multiple LiDAR sensors increases the cost of the system; the technology itself is mature, but not yet optimized for affordable applications. Additionally, LiDAR-generated data is of high volume, making its storage and processing quite difficult, especially for deployment on edge devices. Each frame consists of thousands of points defined by three coordinates in 3D space, and also carries information such as signal intensity, depth, and time of reflection, particularly for applications that rely not just on a single frame but on video sequences.

\subsection{LiDAR Point Cloud Representation}
\textcolor{black}{
A LiDAR scan produces a finite set of 3D points, called a point cloud, and it is represented as follows: \(
\mathcal{P} = \{ p_i \}_{i=1}^N \subset \mathbb{R}^3 \).
Fig.~\ref{cartesian} illustrates a set of points forming the point cloud. Each point is initially represented in spherical (polar) coordinates and then converted to Cartesian coordinates that can be, respectively, expressed as follows:
\[
\begin{aligned}
\text{Polar representation: } & p_i = (\rho_i, \theta_i, \phi_i),\\
\text{Cartesian representation: } & p_i = (x_i, y_i, z_i).
\end{aligned}
\]
\[\text{With:} \hspace{0.2cm}
x_i = \rho_i \cos(\phi_i) \sin(\theta_i), \hspace{0.2cm} y_i = \rho_i \sin(\phi_i) \sin(\theta_i), \hspace{0.2cm} \text{and} \]
\[ z_i = \rho_i \cos(\theta_i).
\]
where $\rho_i$ is the distance to the target, $\phi_i$ is the azimuth angle (horizontal), and $\theta_i$ is the elevation angle (vertical).
Additional attributes such as signal intensity $I_i$ or timestamp~$t_i$ can be attached to each  point from the cloud: \(p_i = (x_i, y_i, z_i, I_i, t_i).
\)
There exist different types of LiDAR based on their scanning range, with some being more common than others, as illustrated in Fig.~\ref{types}. When both $\theta$ and $\phi$ are limited to a specific range, for example $\theta, \phi \in [0, \pi]$, it results in a LiDAR that can scan only the environment facing it, this type is called a front-facing LiDAR. A second type allows a larger perspective by spanning the full azimuth, $\phi \in [0, 2\pi]$, while keeping the elevation $\theta$ limited. As shown in Fig.~\ref{types}, it provides a 360$^\circ$ span on the horizontal plane with a limited elevation angle. Another type allows not only a 360$^\circ$ span in azimuth, but also a wider range in elevation by spanning $\theta$ from $0$ to $\pi/2$, creating a half-spherical field of view.}

\subsection{LiDAR in the Broader Sensing Landscape}
\begin{figure}[t!]
    \centering    \includegraphics[width=0.9\linewidth]{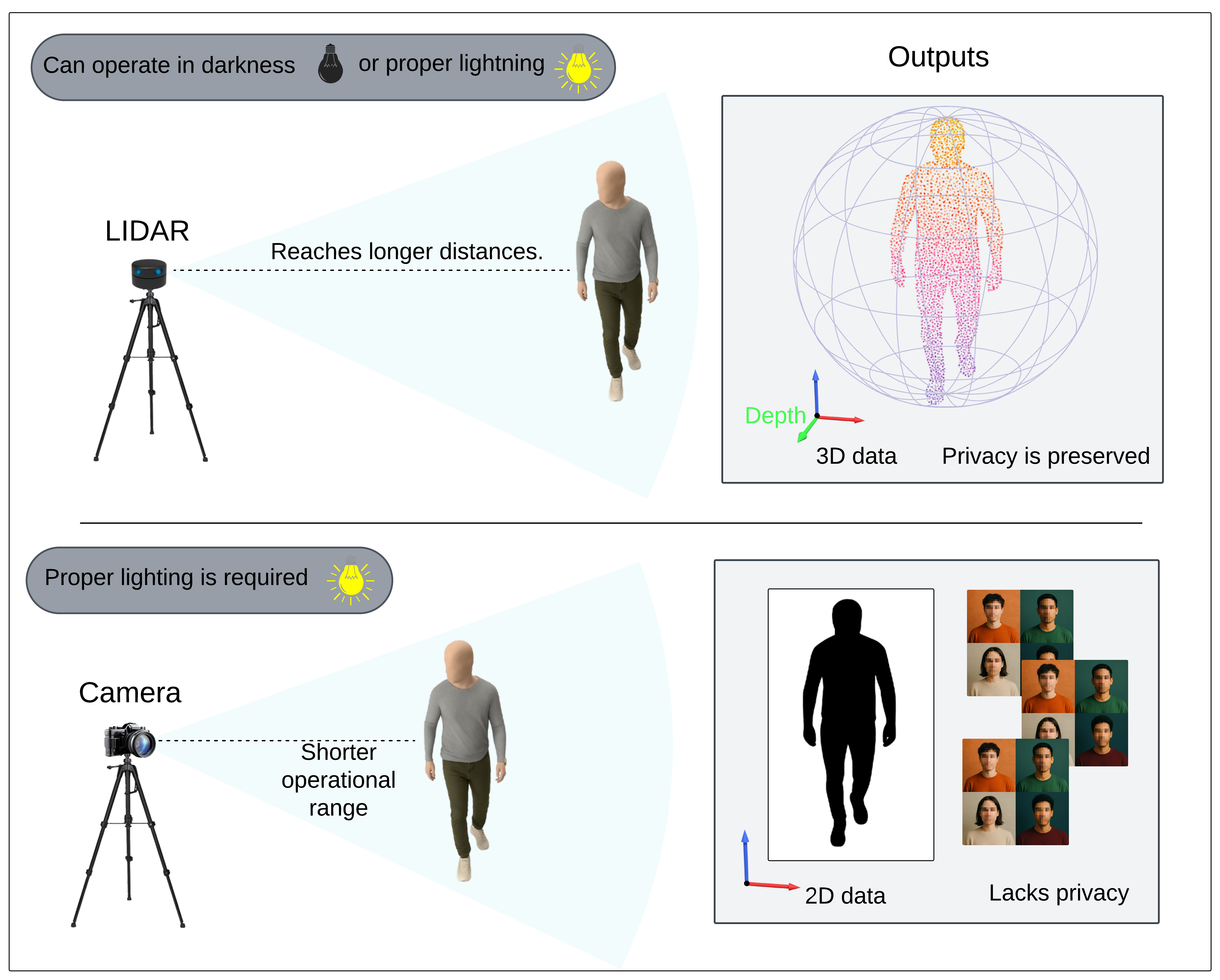}
    \caption{Comparison between LiDAR and camera-based sensing systems: LiDAR (top) operates in both light and dark conditions, offers longer range, outputs 3D depth data, and preserves privacy through anonymized point cloud representation. In contrast, standard cameras (bottom) require adequate lighting, operate at shorter ranges, output 2D data, and may compromise privacy due to identifiable image capture.}
    \label{LiDAR vs camera}
    \vspace{-0.5cm}
\end{figure}

\begin{figure*}[h!]
    \centering \includegraphics[width=0.9\linewidth]{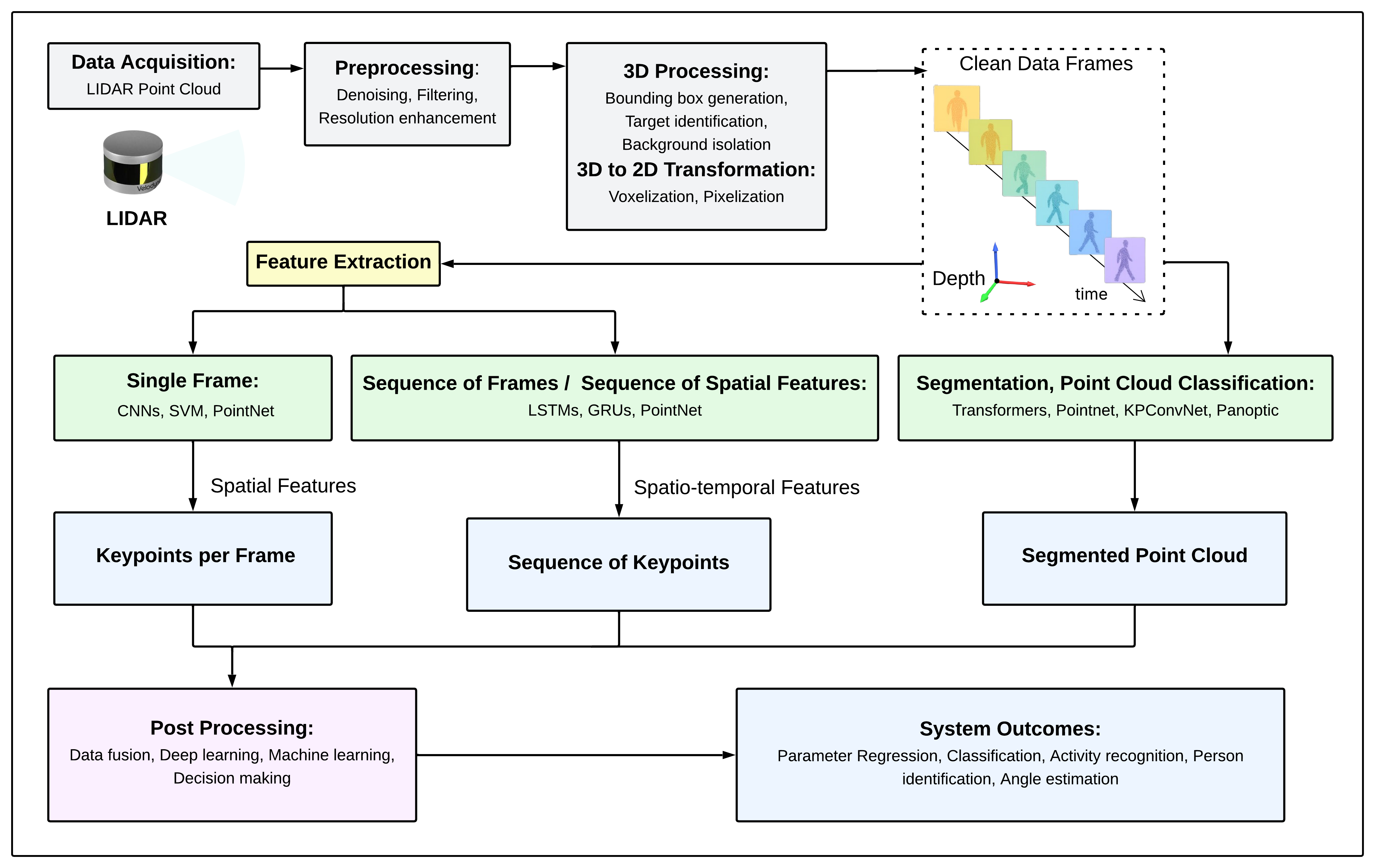}
    \caption{Overall pipeline of LiDAR point cloud processing, including preprocessing, feature extraction, key joint estimation, segmentation, and subsequent tasks such as parameter regression or classification. }
    \label{pipline}
    \vspace{-0.5cm}
\end{figure*}
Fig.~\ref{LiDAR vs camera} illustrates several practical advantages of using LiDAR over camera systems in the context of rehabilitation. \textcolor{black}{LiDAR data preserves individuals’ privacy~\cite{hu2023gait,gaddam2024enhancing}, which is highly recommended in clinical practice for protecting patient confidentiality and dignity, as it does not capture any visually identifiable information about the participants, addressing a major concern in camera-based solutions.} Additionally, it provides richer, multi-dimensional data compared to cameras, which lack depth information, functions well in low light or complete darkness, achieves longer sensing ranges, and provides wider fields of view~\cite{kaewrat2024enhancing,yoon2021development,shen2023lidargait}.

\begin{table*}[t]
\centering

\textcolor{black}{\caption{Comparison of sensing technologies used for rehabilitation across multiple criteria}} 
\renewcommand{\arraystretch}{1.3}
\rowcolors{2}{verylightgray}{white}
\begin{tabular}{p{1.8cm} p{2.4cm} p{2cm} p{2cm} p{2.4cm} p{2.4cm}  p{2.6cm}}
\toprule
\textbf{Criteria} & \textbf{LiDAR} & \textbf{RGB camera} & \textbf{Stereo Camera} & \textbf{IR Camera}  & \textbf{OMC systems}
&\textbf{IMUs} \\
\midrule
\textbf{Raw output data} &
\textbf{3D:} signal intensity, return time, reflectivity &
\textbf{2D:} pixel color values (R, G, B), brightness, saturation &
\textbf{3D:} pixel colors (R, G, B), depth... &
\textbf{2D:} pixel infrared intensity, temperature values& \textbf{3D:} acceleration, velocity, angular velocity &
\textbf{3D:} acceleration, velocity, angular velocity \\

\textbf{Frame rate} & Moderate (5–30 Hz) & High (15–120 Hz) & High (15–120 Hz) & Moderate (5–30 Hz) & High (50–1000 Hz)& High (10–1500 Hz)\\

\textbf{Privacy protection} & High & Low & Low& High & High&  High \\

\textbf{Function in low lighting} & Excellent & Poor & Poor& Excellent & Excellent& Excellent \\

\textbf{Function in high lighting} & Good & Excellent& Excellent & Good &Moderate& Excellent \\

\textbf{Range} & Short to long (5 to 300 m) & Short to medium (1 to 50 m) & Short (1 to 30 m) & Short (1 to 10 m) & Short (1 to 10 m) & Attached to the body  \\ 


\textbf{Setup complexity} & Low & Low & Low & Low &High& Low \\

\textbf{Data volume} & High & Moderate & Moderate &Moderate& Low &Low \\

\textbf{Cost} & Low to moderate (USD 300 to 15 000) & Low (USD 100 to 6 000)   & Low to moderate (USD 200 to 3 000) & Low to moderate (USD 50 to 2 000) & High (USD 10 000 to 50 000)&  Low (USD 1 to 1 000)\\
\bottomrule
\end{tabular}
\label{Comparison}
\footnote{The quantitative information presented in this table is approximate and may vary depending on the provider and the level of system sophistication.}
\vspace{-0.5cm}
\end{table*}

Table~\ref{Comparison} provides a detailed comparison between the different technologies used in the literature for real‑time monitoring of rehabilitation exercises, including LiDAR, various types of cameras, depth sensors, and IMUs. LiDAR provides 3D data along with additional information such as signal intensity, ToF, and material reflectance. It also protects privacy and preserves the dignity of the impaired patient, while allowing fully contactless monitoring, unlike wearable sensors such as IMUs, which need to be attached directly to specific body locations, most often at key joints. 

IMUs can provide many features such as position, velocity, and acceleration in 3D coordinates for each joint. They are low-cost and require a simple setup, but they are vulnerable to errors caused by unwanted movement of the sensors themselves; shifts or displacement of the patches during exercises can falsify the measurements. Depth sensors, on the other hand, also enable contactless monitoring with moderate cost and setup complexity. However, even though they produce depth information as pixel values in a 2D grid, their output remains less detailed and less precise than LiDAR. RGB and stereo cameras are widely used in rehabilitation assessment as well. They have a moderate cost compared to LiDAR and generate much smaller data volumes, which makes the processing lighter. However, they provide no privacy protection, have a shorter operating range, and perform poorly in low‑light conditions. Infrared cameras outperform RGB and stereo cameras in terms of privacy and their ability to function in low‑light environments. \textcolor{black}{Optical motion capture (OMC)} systems use reflective markers attached to key joints and infrared cameras to track their positions. They provide 3D coordinates of the markers and allow the computation of their velocity and acceleration. Although camera-based, these systems offer privacy protection, deliver high frame rate data, and are less affected by ambient lighting due to their reliance on IR illumination. However, they are contact-based, vulnerable to errors caused by marker displacement, and are characterized by high cost and setup complexity.

\subsection{Integration of AI in LiDAR Data Processing}
In the literature and in general, more and more focus is now being directed toward the development of learning-based processing techniques of the LiDAR point cloud. Rule-based processing requires high-resolution data, may fail if points are missing from the cloud, and relies entirely on the determined features, which limits the extracted information to the predefined features, whereas in reality, the sparse and complex data might contain much more information. 

\textcolor{black}{Fig.~\ref{pipline} illustrates the overall pipeline of LiDAR point cloud processing. It begins with raw data in the form of cloud points, which pass through preprocessing steps such as denoising, filtering, and, when needed, 3D-to-2D conversion to adapt the data for 2D models. This is followed by \textcolor{black}{feature extraction}, which can be applied to single frames or sequences of frames to extract key points from the cloud. Alternatively, segmentation can be performed to classify the point clouds into groups. Subsequent tasks include parameter regression, further classification, and decision making to obtain a refined output.}
Afterwards, deep learning models, such as \textcolor{black}{convolutional neural network (CNNs)}, or machine learning models can be used to reduce the dimensionality of point clouds in the first layers of multi-stage pipelines~\cite{wu2024lidar,yamada2020gait,he2023lidar}. The network processes each frame independently, focusing primarily on extracting spatial features. For multi-frame processing, recurrent neural networks such as \textcolor{black}{long-short-term memory (LSTMs) and gated recurrent unit (GRUs}) are employed to capture temporal information encoded in the features of consecutive frames~\cite{gaddam2024enhancing,yamada2020gait,bouazizi2021activity,molano2019robotic,cong2023weakly,ren2024livehps,li2022lidarcap}. Some models require a fixed-size grid as input, which is not naturally satisfied by raw point clouds. Therefore, preprocessing is often needed, for example, using bounding boxes to limit the data boundaries~\cite{ye2024lpformer}, pre-detecting the target such as the presence of a human, or transforming the sparse 3D data into a regular, structured form like 2D pixelized images~\cite{zhang2019environmental} or voxels~\cite{fan2023human}. PointNet,  \textcolor{black}{multilayer perceptron (MLP)-based} architecture specifically designed for 3D point clouds, eliminates the need for such transformations~\cite{qi2017pointnet}.
Moreover, components such as transformer layers, MLP heads, and PointNet based blocks can be trained to perform segmentation~\cite{ma2022research,rinchi2023patients,gaddam2024enhancing}, which involves assigning a label to each point in the cloud and dividing them into specific groups. Examples include semantic segmentation of body parts, such as arm, leg, and head, or scene-oriented segmentation, such as distinguishing background, objects, humans, and others. While the Transformer architecture uses a self-attention mechanism to weigh the importance of different relationships among input points, PointNet relies on shared MLPs and focuses more on global features. PointNet++, an improved version of PointNet, applies the PointNet architecture hierarchically to local neighborhoods, extracting multi-scale information~\cite{qi2017pointnet++}. This hierarchical approach allows it to capture subtle information embedded in the local relationships between points. Similarly, to overcome the limitation of the basic PointNet and its focus on global features, multi-scale sampling can be applied, for example, using KD-Tree neighborhood queries~\cite{ma2022research}. 

Further applications involve key joints estimation, which usually comes in later stages of the pipeline, typically after feature extraction or segmentation. It consists of the determination of $N$ points, each assigned to a specific body joint. These points allow building a skeleton, and with models such as Skinned Multi-Person Linear model (SMPL), it is possible to sketch the skeleton and its movements by providing sequences of the estimated joints~\cite{ren2024livehps,li2022lidarcap}. Keypoint Transformer~\cite{ye2024lpformer}, CoherenceFuse Transformer~\cite{zhang2024neighborhood}, ResNet-based model~\cite{wu2024lidar}, MLPs based model~\cite{li2022lidarcap} can perform joints estimation. This step can be followed by fully connected layers that aim to regress parameters such as movement angles and limb lengths. Further layers can also be added to perform more specific tasks, either for classification or regression, for example, identity recognition, classifying gait as normal or abnormal, or estimating positions such as sitting, standing, or lying down.

Notably, \textcolor{black}{graph convolutional networks (GCNs)} are becoming more widely used with LiDAR data due to their ability to handle the irregular structure of point clouds. They take as input nodes, which represent the points with their 3D coordinates and possibly intensity and other features, along with edges that define the connections between nodes based on their spatial proximity. GCNs can perform classification~\cite{zhao2022safe}, regressing certain parameters, and even predicting key joints~\cite{zhang2024neighborhood,chen2022efficient}.

\section{LiDAR Technology in Rehabilitation applications}

In this section, we investigate the studies that employ LiDAR in ways relevant to the rehabilitation and healthcare of impaired patients. \textcolor{black}{The literature search was conducted using keyword-based queries across major scientific databases, including Google Scholar, IEEE Xplore, Elsevier, and PubMed. The search focused on publications from 2019 to 2025 that involve the use of LiDAR technology as a sensing solution in rehabilitation practices. We focused on this time period because the use of LiDAR has significantly increased in recent years, driven by advancements in sensor optimization and cost reduction, which have made it more accessible~\cite{bi2021survey,saracco2019lidar}. All retrieved references were screened to extract relevant information related to LiDAR applications in rehabilitation and summarized in the following sections.} Our aim is to identify key aspects of each work, including: whether LiDAR was used alone or combined with other devices such as cameras or IMUs; which rehabilitation task was addressed; whether real or synthetic data (or online datasets) were used; whether the solution was AI-based or non-AI, which models were implemented. This distinction is provided to analyze how the complexity and large size of LiDAR data are handled, whether through traditional geometrical methods or learning-based approaches. Key findings and limitations are extracted and we categorize the collected studies into the following four groups: 
\begin{enumerate}[label=(\Alph*)]
\item \textbf{Stationary LiDAR for Human Body Assessment}, where LiDAR is used as a standalone stationary device for tasks such as full-body scanning or gait assessment,
\item \textbf{LiDAR Integrated into Medical Robots}, where it is used either for rehabilitation monitoring or to aid navigation in medical environments,
\item \textbf{Activity Recognition}, which mainly refers to determining the subject’s position or recognizing gestures; and
\item \textbf{Other Applications}, covering remaining innovative or less common uses of LiDAR for other rehabilitation tasks, 
\end{enumerate}
and we provide a detailed overview of the different systems and techniques used in each category.
\subsection{Stationary LiDAR for Human Body Assessment}
One or more LiDAR units can function as standalone systems to capture point cloud data, enabling the scanning of specific limbs or the entire body, and supporting gait analysis in patients.
\subsubsection{Body Part Scanning}
To support detailed analysis of anatomical structures, medical staff often rely on 2D or 3D body scans. These scans are also useful for tracking rehabilitation progress and maintaining patient records. With recent advancements, 3D models are most commonly generated using LiDAR, which has demonstrated high precision in modeling body parts. For example, back scans can provide precise details such as vertebral deviation, the angle and depth of kyphosis and lordosis, as well as spinal alignment in all three anatomical planes: sagittal, coronal, and transverse~\cite{febbi2024acute}.

LiDAR-based approaches have shown improved accuracy compared to image-based scans~\cite{klein2024angle,barzegar2024joint}, and have achieved satisfactory results in clinical use~\cite{paoli2020sensor} and laboratory trials~\cite{quagliato2025optimizing}. For instance, an error of -0.5\% was reported when comparing the measurement outcomes from a LiDAR scan to those obtained using conventional marker-based techniques~\cite{Mcconnochie2025LIDAR}. In this study, the point cloud data were processed using SMPL model, and showed a 0.96 correlation between repeated measurements. Similarly, in another study~\cite{oberhofer2024feasibility}, a 0.90 agreement with marker-based measurements was reported, specifically for thigh measurements. In this work, Oberhofer et al. used the Polycam app~\cite{polycam2025} and emphasized the importance of performing repeated scans and averaging the results, which was shown to improve overall scan quality. Reimer et al., in their study~\cite{reimer2022evaluating}, used Apple’s ARKit 3D LiDAR~\cite{applearkit2025} and compared its measurements to those from the gold-standard, camera-based Vicon system~\cite{vicon2025}. The ARKit showed a mean squared error of 18.8 $\pm$ 12.12°, and a Spearman correlation coefficient of 0.76 when compared to Vicon measurements. Furthermore, De Sire et al. used the Occipital 3D LiDAR along with Captevia Rodin4D software~\cite{captevia2025} to perform precise upper limb scans, and statistical analysis showed a 0.99 correlation when compared to the circumferential method~\cite{de2020three}. Moreover, an example of lower limb rehabilitation is the treatment of clubfoot patients, a disorder that can be acquired after stroke, brain injury, or be present from birth. In some cases, physical therapy plays a crucial role in helping the foot muscles regain their strength, as these patients often stand on their ankles. Lay-Ekuakille et al. proposed the use of LiDAR and geometry-based algorithms to extract foot measurements~\cite{lay2023preventing}. Results from two healthy participants showed that all measurements fell within the expected range of a normal foot, highlighting the future potential of LiDAR techniques to classify the severity of clubfoot in patients.

In more specific use cases, LiDAR can serve as a modeling tool to fabricate personalized paramedical devices. Personalization of artificial limbs for amputees, which are attached via sockets, plays a crucial role in the functionality of the prosthesis. Using standard silicone sockets may not suit every case, especially since the size of the residual limb can vary temporarily when standing or sitting. Researchers in~\cite{squibb2024high} used LiDAR to scan the residual limb of transtibial amputees to design better fitting sockets (see Fig.~\ref{amputee}). Their results showed a measurement error of 0.17\% on a known cylindrical shape and 0.7\% error on a real amputee limb, equivalent to 8.1~ml in volume. They also highlighted their system’s ability to capture the temporary changes in dimensions, with a 4\% change corresponding to a 46~ml volume difference observed. Customization also enhances stability in exoskeletons, as highlighted by Secciani et al., who fabricated customized hand splints for their hand exoskeleton using 3D LiDAR, the generate point clouds were processed using PolyWorks and Geomagic Design X software~\cite{secciani2021wearable}.\\
\begin{figure}[t!]
    \centering
    \includegraphics[width=0.9\linewidth]{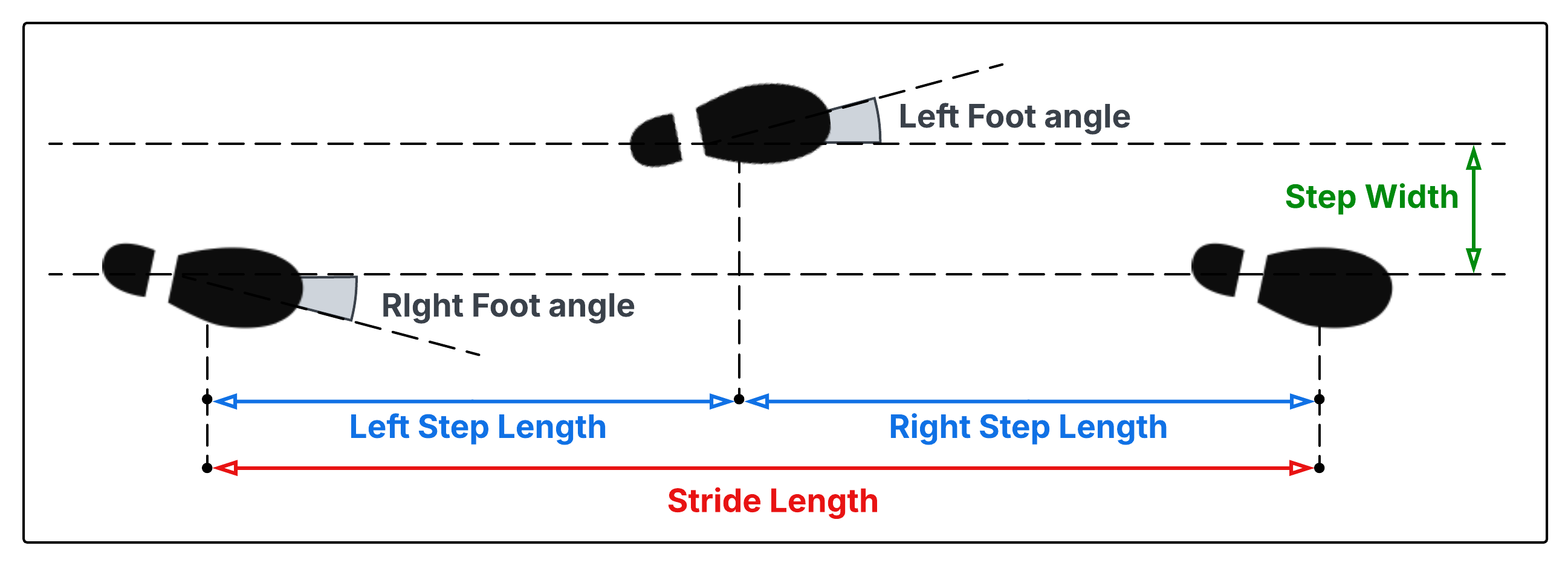}
    \caption{ Key gait parameters including step width, stride length, step length, and foot angles.}
    \label{gait-path}
    \vspace{-0.5cm}

\end{figure}

\begin{figure*}[h!]
 \centering
 \includegraphics[width=0.8\linewidth]{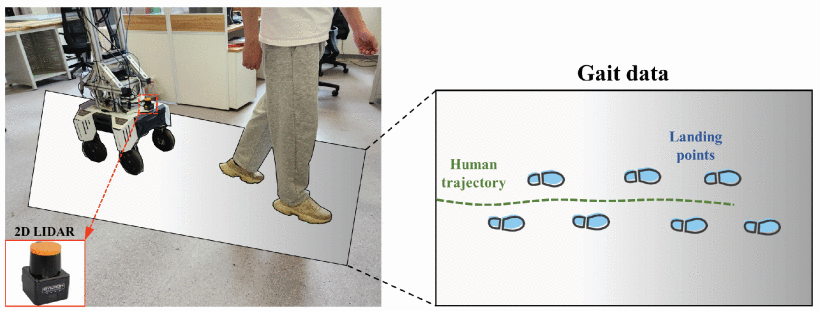}
    \caption{LiDAR sensor mounted on a mobile robot that follows the patient’s trajectory while analyzing gait patterns~\cite{tan2023quantitative}.}
    \label{rob-path}
    \vspace{-0.5cm}
\end{figure*}

\subsubsection{Gait Analysis}
Gait refers to the manner in which a person walks characterized by a set of temporal and spatial parameters (see Fig.~\ref{gait-path}), 
Botros, Single et al., in their studies~\cite{botros2021contactless, single2024transferable}, used 2D LiDARs (UST--LX-H01)~\cite{hokuyoUST10LXH012025} and geometry-based algorithms to estimate step and stride length and time, velocity, and cadence across different walking patterns. These gait parameters are mathematically related as follows~\cite{carollo2015quantitative}:
\[
v = \frac{C \cdot L_{stride}}{120} \quad (\text{m/s}), \hspace{0.2cm}
L_{stride} = \frac{v \cdot 120}{C} \quad (\text{m})
\]
\[
L_{step} = \frac{v \cdot 60}{C} \quad (\text{m}), \hspace{0.2cm}
L_{stride} = 2 \cdot L_{step}
\]
Where $v$ is the walking speed, $C$ is the cadence, representing the number of steps per minute, $L_{stride}$ is the stride length, and $L_{step}$ is the step length.  
Botros, Single et al. reported correlations with ground truth measurements, such as those from IMUs, higher than 0.94 for all parameters, except for cadence, which showed a correlation of 0.91. Botros et al.’s system also demonstrated stability over time, maintaining performance when tested continuously for 12 hours~\cite{botros2021contactless}.

Gait assessment using LiDAR showed medically valid performance with less than 8\% error~\cite{zheng2023gait}, and improved accuracy compared to image-based techniques. In~\cite{yoon2021development}, the RPLiDAR A3M1~\cite{rplidarA3M12025} data processed with the Inertia-based Object Tracking Algorithm achieved 0.95 correlation with the gold standard Raptor-E infrared camera system~\cite{raptorSystem2025}. The error was also reduced from $116.3\,\pm 69.6$~mm using the RGB-based OpenVINO toolkit~\cite{openvino2025} to $46.2\,\pm 17.8$~mm. The authors observed that the naturally slow walking patterns of elderly individuals might help improve accuracy. Moreover, Duong et al.\ experimented with two LiDAR placements: parallel and horizontal to the walking path~\cite{duong2020human}. Using clustering and geometry-based methods, they first separated the left and right legs, then estimated the walking parameters. Interpolation was applied when a leg was not detected. For parallel placement, the errors were 5.9~cm and 4.3~cm; for horizontal placement, 4.3~cm and 3~cm, for step length and walking distance, respectively~\cite{duong2020human}.
Given the nature of gait analysis, which relies not only on spatial but also on temporal information, researchers define customized pipeline based on models such as CNNs, transformers, LSTMs, or \textcolor{black}{graph neural network (GNNs)}~\cite{gaddam2024enhancing, zhang5829380gaitcloud+, shen2025lidargait++}, or by utilizing available pretrained models and software solutions~\cite{kaewrat2024enhancing}. For instance, Farewik et al.\ improved the correlation between the Apple ARKit 3~\cite{applearkit2025} system’s estimations and ground truth from 0.51–0.83 to 0.96–0.98 by using a nonlinear neural network regression model~\cite{farewik2022markerless}. In another study~\cite{hu2023gait}, researchers performed geometric alignment of two LiDAR point clouds, followed by density-based spatial clustering and Extended Kalman Filter to identify and track the foot's position. They extracted parameters such as step length, step width, and gait speed, and they were able to define useful thresholds for detecting gait impairment: if variability in step width, gait speed, and step length exceeds 0.15, 0.25, and 0.3 respectively, the subject can be considered gait impaired. 
The walking pattern, as reported by some studies, consists of two phases: a dynamic phase at the beginning of the walk, where parameters such as speed increase, followed by a static phase where the speed plateaus. It is important to remove point cloud frames of the dynamic phase before performing parameter estimation. For elderly individuals, it typically takes about $0.92 \pm 0.51\,\text{m}$ to reach the static phase~\cite{ji2024comparative}. The study also reports a 33\% correction in the estimation of gait speed when removing the dynamic phase from the recording~\cite{ji2024comparative}.
\subsection{LiDAR Integrated into Medical Robots}
In this section, we present a range of studies that integrate LiDAR into rehabilitation robots. These studies can be categorized into two main groups. The first involves the use of LiDAR to assess the rehabilitation process by tracking patient progress through body part scanning. The second category focuses on the use of LiDAR for environment mapping and obstacle detection, acting as assistive systems. 
\textcolor{black}{
\subsubsection{Gait Rehabilitation Monitoring}
The integration of LiDAR in rehabilitation robots and exoskeletons enables real-time assessment of gait parameters. The progress of patients undergoing rehabilitation can be reflected in specific measurements, such as joint angles, range of motion, and the timing of particular movements. Quantitative analysis of these parameters allows for an objective evaluation of recovery. Robotic systems can directly capture these indicators from a close perspective that is difficult for caregivers to achieve.} The use of rehabilitation robots has also been shown to improve patient safety~\cite{yao2024advancements, chadha2025omni}, as they provide accurate real-time feedback and smooth human–robot interaction, enabling the prediction of potential falls or dangerous situations and helping to prevent injuries during rehabilitation exercises. Fig.~\ref{rob-path} shows a 2D LiDAR mounted on a robot that follows a patient’s trajectory while analyzing his gait. A LiDAR-based walker robot proposed by Sakdarakse et al. followed the subject's lower limb in real time during gait training, and achieved an estimation error of 6.42\% for step length and 1.95\% for stride length~\cite{sakdarakse2020development}. Lee et al. tested an indoor accompaniment dog robot on stroke patients undergoing gait rehabilitation; the robot, using 2D LiDAR, maintained a certain distance from the patient, ensuring reachability when needed~\cite{lee2020development}. Furthermore, tests on five healthy participants using a 2D LiDAR-based walker robot showed reliable performance in real-time gait tracking and walker control~\cite{vithanage2024smart}. In another study involving eight Parkinson’s disease patients with gait impairment~\cite{tan2023quantitative}, the UST-10LX 2D LiDAR~\cite{hokuyoUST10LXH012025} was used alongside Adaptive density-based spatial clustering of Applications with Noise (DBSCAN), intersection-over-union-based clustering, a Kalman filter, polynomial regression, and the proposed gait evaluation function. The system achieved an error of less than 0.05~m and an F1-score of 0.98~\cite{tan2023quantitative}. 
The F1-score is computed as the harmonic mean of Precision and Recall, which are defined as:
\[\text{Precision} = \frac{\text{True Positives}}{\text{True Positives} + \text{False Positives}}, \]
\[\text{Recall} = \frac{\text{True Positives}}{\text{True Positives} + \text{False Negatives}}, \]
\[\text{F1-score} = 2 \cdot \frac{\text{Precision} \cdot \text{Recall}}{\text{Precision} + \text{Recall}} .\]
True Positives correspond to correctly detected gait abnormalities, False Positives are instances incorrectly identified as abnormal, and False Negatives are missed abnormalities.
Moreover, intelligent robots can provide advanced safety features, For example, researchers in~\cite{zhao2020smart} suggested the detection of human action intention and emergency cases, such as falls based on neural networks, as well as in ~\cite{zhao2022safe}, who proposed the detection of gait abnormalities using a Fuzzy Petri Net Node Interference, achieving an accuracy of 91.2\%. 
\subsubsection{LiDAR-based navigation assistive systems}
\begin{figure}[h!]
 \centering
\includegraphics[width=0.9\linewidth]{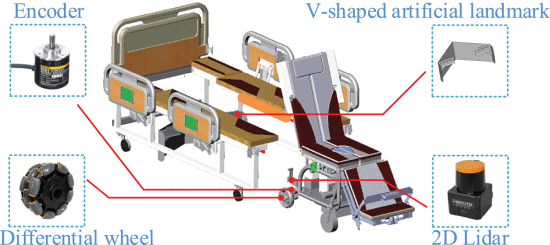}
    \caption{Automated Wheelchair Docking to a Body-Separated Nursing Bed based on 2D LiDAR Sensing~\cite{zhu2021wheelchair}.}
    \label{wheelchair}
\end{figure}

The independence that assistive \textcolor{black}{robots and devices} provide to patients and individuals with impairments has led to more demand for their use. Recent studies focus on integrating new features in these \textcolor{black}{systems} to handle any situation that the patient can face during navigation, for full independency. For example, Kim et al. developed a 2D LiDAR-based elevator detection system that is able to detect up to 80\% of elevators’ presence from the side view, which is usually challenging, and localize them for smooth access with a 6.11~cm error~\cite{kim2023study,kim2023A}. Zhu et al. proposed a novel LiDAR-based docking system between wheelchair and nursing bed, which allows patients to align correctly with the bed to safely be transferred to it (see Fig.~\ref{wheelchair}), the system achieved an error of less than 0.02~m and 2.5° deviation~\cite{zhu2021wheelchair}.
Another situation that a patient can face is staircases passage, specifically for exoskeleton wearers; the modeling of the staircase is of importance to allow safe ascending or descending. It can be performed using 1D LiDAR and from different points of view (facing or not facing), where a 3~cm and 3° tolerance is enough for safe navigation~\cite{raineri2021real}.

Further, researchers in~\cite{rufei2022research} underscored the risks related to floor inconsistencies such as cracks and holes, which pose a high risk of injury to wheelchair users. They proposed the use of 3D LiDAR and geometry-based algorithms, which outperformed image-based pothole detection models by correctly detecting 31 out of 34 potholes compared to 29~\cite{rufei2022research}.
Obstacle avoidance either relies on LiDAR-based geometrical operations to detect closeness to obstacles~\cite{lee2022hardware,ton2018LIDAR} or multi-modal fused algorithms. For example, Su et al. achieved robust performance using 2D LiDAR, odometry and IMU through Extended Kalman Filter~\cite{su2021research}, or Molano et al., who proposed a system that fuses LiDAR, a depth camera, and a pressure sensor to detect static obstacles using LSTM blocks~\cite{molano2019robotic}.
LiDAR data is usually used for SLAM, which stands for \textcolor{black}{\textit{simultaneous localization and mapping}}. The robot senses the environment via the LiDAR and detects landmarks such as walls and obstacles, allowing it to map the environment and localize itself at any time~\cite{su2021research, xu2025conav, mane2025mark}. Slade et al. proposed a LiDAR-based SLAM navigation cane for vision-impaired patients that improved their navigation by $35 \pm 12\%$. The cane is affordable, says Slade et al., and can map the environment in real time and provide sound feedback to properly guide patients~\cite{slade2021multimodal}. SLAM was also utilized by Hakkim et al., who in their study~\cite{hakkim2024LIDAR} proposed a prototype for the use of YDLIDAR-X4 2D LiDAR~\cite{ydlidarX42025} in medicine delivery robots inside hospitals. This application requires high-precision mapping and accurate obstacle detection to enable smooth and safe delivery in such actively dynamic environments.
Overall, the integration of LiDAR in assistive \textcolor{black}{systems} has shown great performance in complex medical environments~\cite{ibrayev2024development,eun2023self}, in home rehabilitation~\cite{mansoubi2024investigation}, and potential for operation in narrow dense spaces~\cite{szaj2021mechatronic}. Researchers also highlight the importance of enabling intelligent AI-driven decisions while still allowing human commands when needed, and safety should remain the top priority when conflicts arise~\cite{sun2024review,szaj2021mechatronic}. For example, the Temporal-Difference learning with General Value Functions, based on a reinforcement learning system proposed by Faridi et al., was able to reduce user walking mode switches in a lower limb exoskeleton by 42.44\% for safer navigation, by automatically predicting switching needs based on LiDAR scanning~\cite{faridi2022machine}.

\subsection{LiDAR-based Human Activity Recognition}
\begin{figure}[t!]
 \centering
 \includegraphics[width=0.9\linewidth]{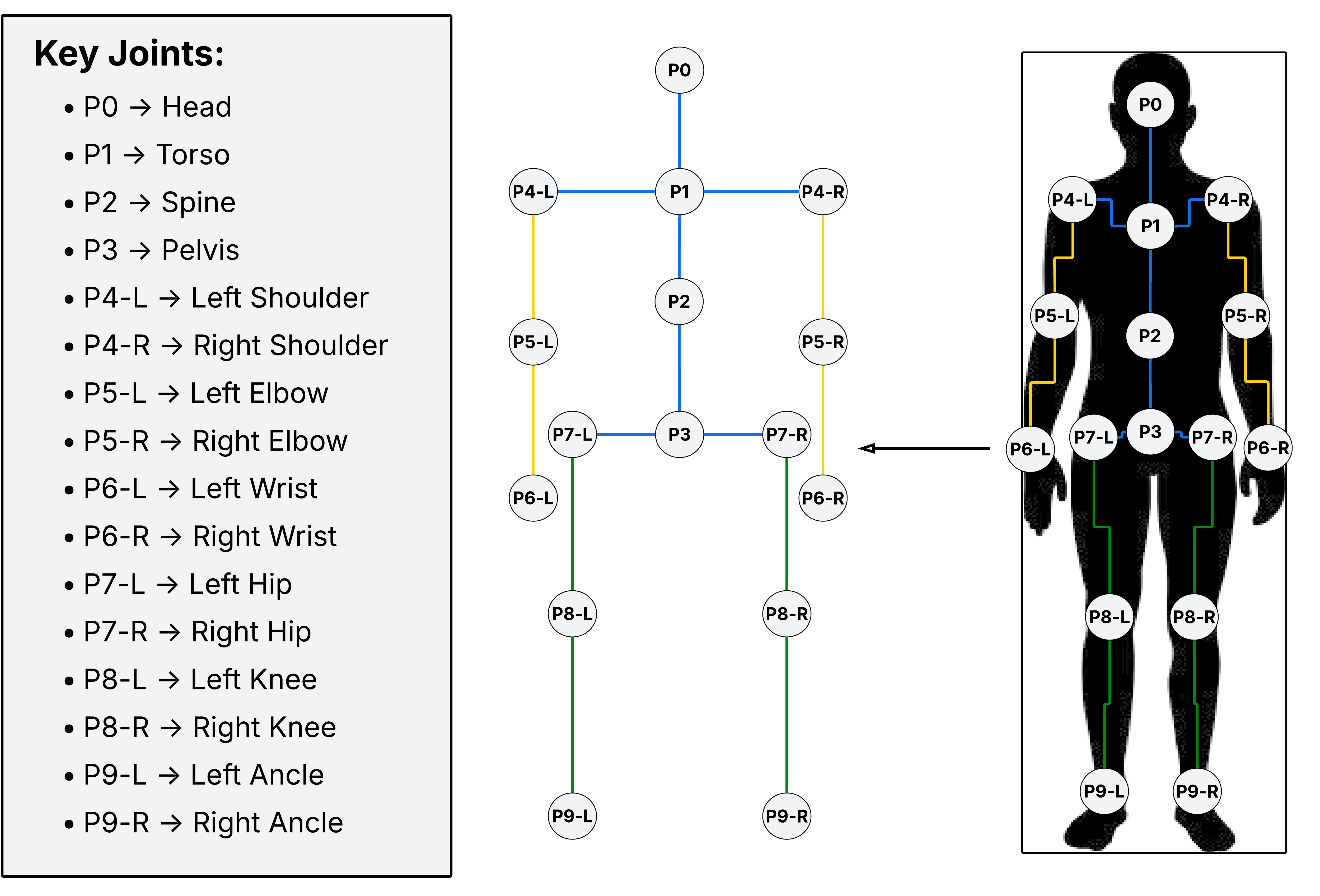}
    \caption{Generated human skeleton using 3D LiDAR point cloud data. The left panel shows the body segments and joints, the center shows how they are connected, and the right displays the skeleton on top of the LiDAR data.}
    \vspace{-0.5cm}
    \label{skeleton}
\end{figure}

Position and activity recognition can find applications in hospitals, helping caregivers monitor patients while they are in their rooms, during rehabilitation, or at home~\cite{10145011}. The first step after collecting the point clouds is to either classify these points in order to segment the body, or to detect key joints, reducing the dimensionality. Then, based on these segments or key points, 2D or 3D skeletons are reconstructed, and the position can be estimated. Fig.~\ref{skeleton} shows the key segments and joints of the human body, which, when connected together, form the skeletal structure.

LiDAR is widely used for this application; Patil et al.~\cite{patil2020fusion} proposed a real-time pose estimation method where point clouds were used for skeleton construction and then fused with IMU data. They achieved a 5° error in joint angle prediction, $\pm 3$~cm in height estimation, and the pelvis tracking accuracy was within $\pm 5$~cm. Wu et al.~\cite{wu2024lidar} used the L515 3D LiDAR~\cite{intelL5152025} to gather point clouds, which were mapped into depth images, then processed with an improved Anchor-to-Joint regression network based on ResNet50 to extract 15 key skeleton points, followed by classification using \textcolor{black}{support vector machine (SVM)} model, achieving an accuracy of 93.46\% and a recognition speed of 42 frames per second. Moreover, the researchers in~\cite{bouazizi2021activity} proposed an in-home surveillance system designed for the elderly that detects daily activities with an accuracy of 92.3\%. The system also identifies gait abnormalities related to dizziness and falls by using LSTM blocks. It can monitor up to three individuals simultaneously, enabling timely assistance when a fall occurs. Zhao et al.~\cite{zhao2024lidar} proposed the use of Transformer-based MotionBERT3D on data collected from the Livox Mid-70 3D LiDAR~\cite{livoxMid702025} to classify participants’ positions into standing, sitting, walking, squatting, and lying, outperforming camera-based models. The authors highlighted the advantage of LiDAR in preserving participants’ privacy. Further, a two-stage model was proposed in~\cite{furst2021hperl}, using a feature extractor based on Region Proposal Network and Aggregate View Object Detection, followed by a pose estimator LCRNet~\cite{rogez2017lcrnet}. They achieved 70.22\% correct key point estimation compared to 65.92\% when relying on RGB images only. Additionally, the center point depth error dropped from 4.88~mm to 0.95~mm, and the \textcolor{black}{mean per joint position error (MPJPE)} was reduced by a factor of 1.9.
\textcolor{black}{
The MPJPE is a common metric in 3D human pose estimation. It measures the average Euclidean distance between predicted and ground-truth joint positions. For a single frame with $J$ key joints, let:
\begin{itemize}
    \item $\mathbf{X}_j \in \mathbb{R}^3$ represents the ground-truth position of joint $j$,
    \item $\hat{\mathbf{X}}_j \in \mathbb{R}^3$ represents the predicted position of joint $j$.
\end{itemize}
The per-joint Euclidean error is:
\[
e_j = \|\hat{\mathbf{X}}_j - \mathbf{X}_j\|_2.
\]
The MPJPE is then defined as the mean over all joints:
\[
\text{MPJPE} = \frac{1}{J} \sum_{j=1}^{J} \|\hat{\mathbf{X}}_j - \mathbf{X}_j\|_2.
\]
For multi frames, let $F$ be the number of frames, the MPJPE is the averaged MPJPE over all frames:
\[
\text{MPJPE} = \frac{1}{F} \sum_{f=1}^{F} \frac{1}{J} \sum_{j=1}^{J} \|\hat{\mathbf{X}}_j^{(f)} - \mathbf{X}_j^{(f)}\|_2,
\]
where $\mathbf{X}_j^{(f)}$ and $\hat{\mathbf{X}}_j^{(f)}$ are the ground-truth and predicted positions of joint $j$ in frame $f$.
}

Ye et al.~\cite{ye2024lpformer} proposed the LPFormer model, also a two-stage approach, based on LIDARMultiNet, which takes point clouds as input and generates 3D bounding boxes and semantic segmentation. The second stage is a Transformer-based network that predicts key joint points from the first-stage output. This model surpassed different state-of-the-art techniques, with a 34\% enhancement in accurate 3D joint position prediction and a 70\% improvement in MPJPE.

Zhang et al.~\cite{zhang2024neighborhood} suggested the integration of background point clouds for real-time position tracking. They reported a 7.08~mm reduction on average MPJPE and outperformed the P4Transformer model~\cite{wen2022point} on the LIDARHuman26M dataset~\cite{li2022lidarcap}, achieving 95.79\% accuracy compared to 85.64\%. Another study~\cite{patil2021open} fused LiDAR data with IMU, with the aim of reducing the necessity of two LiDARs in such applications. They reported an error of 3~cm with a single LiDAR compared to up to 5~cm when using two LiDARs due to calibration errors. Additionally, their technique showed a reduction in root mean square error by a factor of 3, compared to when relying solely on IMU.
Bauer et al.~\cite{bauer2023weakly}, on the other hand, showed improved performance when combining LiDAR data with a monocular camera. They reported an MPJPE of 12.52 for camera only, 11.22 for LiDAR only, and 8.58 when merging both sensing techniques on the Waymo dataset~\cite{sun2020scalability}. They also suggested the use of a weakly supervised model, eliminating the need for pre-labeled 2D or 3D key points as ground truth. Pseudo-labels are created by projecting point clouds onto the images and selecting nearby points. They used a Lifting network to process the 2D key points and a PointNet-based network for the point clouds, then fused both via dense layers to predict 3D positions.
Zanfir et al.~\cite{zan} also highlight the advantages of fusing LiDAR with camera data for better accuracy. They used a U-Net model for feature extraction from images and Random Fourier embedding for point cloud data, and proposed the HUM3DIL model, based on a transformer encoder and MLP, outperforming with a 6.72~cm MPJPE, compared to image-based models: ContextPose~\cite{ma2021context}, a multi-modal model~\cite{zheng2022multi}, and Thundr~\cite{zanfir2021thundr}, which achieved 10.82, 10.32, and 9.62~cm respectively. Similarly, Fan et al.~\cite{fan2023human} reported a 24\% decrease in MPJPE when merging camera and LiDAR data using the Multi-Modal-VoxelPos approach. Li et al.~\cite{li2022lidarcap} proposed another multi-modal solution combining LiDAR, camera, and IMU data, and introduced the LIDARCap model, which uses PointNet++ as a feature encoder, an inverse kinematics solver, and an SMPL optimizer for position regression. They reported a percentage of correct key position estimation within 30\% tolerance of 86\%, compared to 49\% achieved by the human mesh recovery image-based framework~\cite{kanazawa2018end}.
\begin{figure}[t!]
    \centering    \includegraphics[width=1\linewidth]{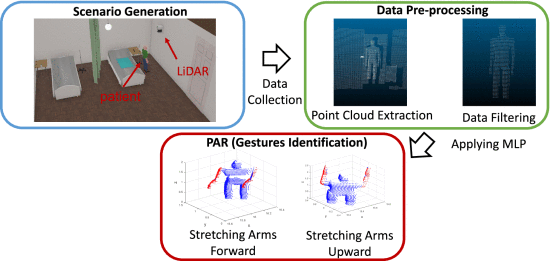}
    \caption{Overview of the LiDAR-based patient activity recognition framework~\cite{rinchi2023patients}, including scenario generation, 3D point cloud data collection and preprocessing, and gesture classification using a MLP-based model. Example gestures include arm stretching movements in different directions.}
    \label{gesture}
    \vspace{-0.5cm}

\end{figure}
A more specific application of position estimation focuses on the upper limb, such as gesture identification, which can be useful for patients who have difficulty speaking. Rinchi et al.~\cite{rinchi2023patients} used an MLP-based model as shown in Fig.~\ref{gesture}, successfully achieving up to 92.5\% accuracy in classifying different hand positions, such as stretching forward, upward, downward at 45°, and others. Similarly, aiming to classify hand gestures such as hand opening, wrist extension and flexion, and fist clenching, He et al. conducted experiments on both healthy and stroke patients. Their SVM classifier achieved 92\% accuracy in real-time performance~\cite{he2023lidar}.

\subsection{Other Applications}
LiDAR technology also shows potential in applications beyond rehabilitation assessment, body scanning, and gait analysis. Song et al.~\cite{song2023smartphone} proposed using LiDAR scanning to assess the size of flat wounds, achieving a correlation of 0.99 with ground truth ruler measurements. This method enables non-contact wound size measurement, reducing the risk of contamination. LiDAR has also been applied as a quality control tool in the fabrication of medical devices. Tian et al.~\cite{tian2023compact} used it during the manufacturing of a waist rehabilitation chair with pillow support to assess the accuracy of the lumbar curve fitting, while Secciani et al.~\cite{secciani2021wearable} employed it to fabricate customized hand splints for upper limb exoskeletons, enhancing stability.

\textcolor{black}{Another promising use is person identification based on gait patterns~\cite{hao2025horgait, zhang2025gaitcloud, an20253d}. In hospital environments, it can support patient monitoring, ensure correct identification without intrusive methods.} This application aims to identify individuals through their gait, scanned by LiDAR as a biometric identity comparable to fingerprints or facial recognition. Han et al.~\cite{han2024gait} extracted a set of geometric and dynamic temporal features from point clouds and applied CNNs based on ResNet architecture combined with MLPs, improving identification accuracy by 3.42\% and 6.61\% on the SUSTech1K~\cite{shen2023lidargait} and FreeGait~\cite{han2024gait} datasets, respectively, compared to other state-of-the-art models. Similarly, Shen et al.~\cite{shen2023lidargait} achieved competitive results with a CNN-based model, specifically GaitBase~\cite{fan2023opengait}, originally designed for images and adapted for point cloud data. Yamada et al.~\cite{yamada2020gait} also used CNN-based encoders along with LSTMs for classification, achieving around 60\% accuracy. They highlighted the limitations of RGB-based cameras in outdoor and long-range scenarios and observed that leg movement plays the most significant role in correct classification, while arm movements, such as crossing or touching the face, tend to reduce identification accuracy.

A further potential use is in upper limb exoskeletons as a visualization tool~\cite{cheng2024efficient}, often combined with cameras that provide 2D information, while LiDAR adds depth. For example, the Velodyne-16 LiDAR was integrated with a camera into a teleoperation exoskeleton system, where the two sensors together provided visual feedback to the user, enabling remote operation via the exoskeleton. This system achieved 100\% success when tested by both beginners and experienced users on highly precise tasks, such as inserting a pin into a hole with only a 0.02~mm diameter difference. 

Another example uses LiDAR not only as a visualization tool but also to determine the shapes and sizes of objects, assisting in commanding a prosthetic hand for smooth grasping~\cite{cui2024research}. The pipeline employed YOLOv8 to process the camera data for object pre-classification, while DBSCAN and DWT were applied to the LiDAR data, followed by fusion of both outputs. This system achieved 96\% accuracy in shape recognition and 91\% success in grasping, with an error range of 5–11~mm.
\\

Overall, the innovative integration of LiDAR into rehabilitation applications, post-injury care, and hospital use has demonstrated competitive performance, offering many advantages over other techniques. It achieves high accuracy in classification, regression, and segmentation tasks, while also enabling long-range, remote applications and ensuring privacy protection. These strengths make LiDAR an ideal solution for rehabilitation. Its potential spans static and dynamic scanning, full- or partial-body focus, use in assistive device navigation for individuals with impairments, integration into real-time robotic systems to closely follow therapy exercises, activity and posture recognition, and many other applications.

\section{Literature Analysis and Takeaways}
In this section, we perform a statistical analysis of the examined studies and summarize the main insights obtained through our comprehensive literature review.

\begin{figure}[t!]
    \centering
\includegraphics[width=0.9\linewidth]{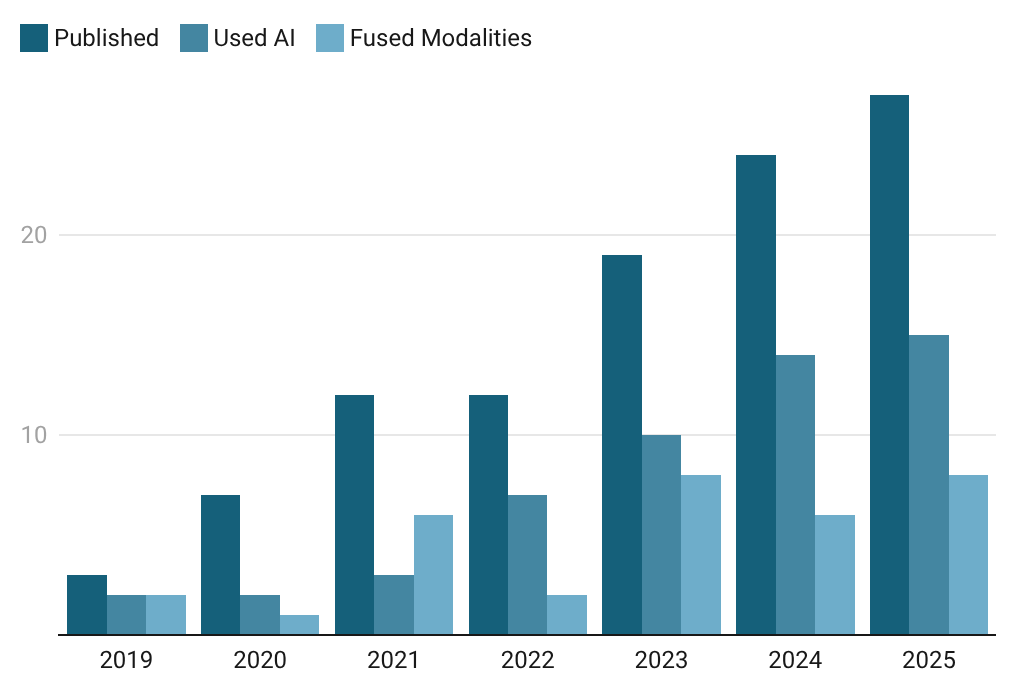}
    \caption{Statistical trends in LiDAR-based healthcare research (2019–2024): Number of studies published, studies applying AI, and studies incorporating modality fusion with LiDAR.}
    \label{graph1}
        \vspace{-0.5cm}

\end{figure}
\textbf{$\bullet$ Growing interest in the field:} We focus our research on the time period from 2019 to 2025. In Fig.~\ref{graph1}, we investigate the general interest of researchers and practitioners in exploring the use of LiDAR for rehabilitation tasks. By the year 2024, the number of references found was almost eight times higher than in 2019, demonstrating that researchers are becoming more interested in the field and more innovative in the use of LiDAR in healthcare applications.

\textbf{$\bullet$ AI becoming increasingly popular:} Fig.~\ref{graph1} also explores the level of AI integration in the developed LiDAR-based solution. \textcolor{black}{It indicates} a clear increase in works that involve AI tools to process LiDAR point clouds. In recent years (2022, 2023, 2024, and 2025), more than half of the references we found present AI-based solutions, showing satisfactory performance. Given this trend, we suggest that AI will become even more predominant in this field, surpassing the reliance on purely geometrical and mathematically deterministic models. The reason is that, in cases of noisy data, occlusion, or interference, key points from the point cloud might be missing and these models tend to fail. In contrast, AI-based solutions can compensate for missing data by leveraging other features to successfully complete the task.

\begin{table}[t]
\begin{center}
\caption{\label{tab_ai_summary}Summary of Research Studies Utilizing LiDAR and AI Techniques for Human-Centered Perception Tasks}
\vspace{-0.5cm}
\begin{tabular}{|c|c|c|c|}
\hline
Ref & Year & AI Tasks & AI Techniques \\
\hline
\cite{yamada2020gait} & 2020 & Gait recognition & CNN + LSTM \\
\cite{zhao2020smart} & 2020 & Gait recognition & MLP \\
\cite{zheng2021multi} & 2021 & Pose estimation & PointNet + MLP \\
\cite{furst2021hperl} & 2021 & Pose estimation & RPN (AVOD) \\
\cite{faridi2022machine} & 2022 & Walking mode & TD + GVF \\
\cite{bouazizi2021activity} & 2022 & Action recognition & CNN + LSTM \\
\cite{zhao2022safe} & 2022 & Gait recognition & NIFPN \\
\cite{ma2022research} & 2022 & Body segmentation & PointNet + Transformers \\
\cite{li2022lidarcap} & 2022 & Pose estimation & PointNet + MLP \\
\cite{rinchi2023patients} & 2023 & Arms segmentation & MLP \\
\cite{fan2023human} & 2023 & Pose estimation & CNN + MLP \\
\cite{shen2023lidargait} & 2023 & Gait recognition & CNN + MLP \\
\cite{he2023lidar} & 2023 & Action recognition & SVM \\
\cite{zan} & 2023 & Pose estimation & RFE + Transformers + MLP \\
\cite{cong2023weakly} & 2023 & Pose estimation & PointNet + Transformers + MLP \\
\cite{ren2024livehps} & 2024 & Pose estimation & PointNet + MLP \\
\cite{wu2024lidar} & 2024 & Action recognition & CNN + SVM \\
\cite{wu2024lidar} & 2024 & Pose estimation & CNN + SVM \\
\cite{zhang2024neighborhood} & 2024 & Pose estimation & PointNet + Transformers \\
\cite{han2024gait} & 2024 & Gait recognition & CNN + MLP \\
\cite{kovacs2024lidpose} & 2024 & Pose estimation & Transformers + MLP \\
\cite{ye2024lpformer} & 2024 & Pose estimation & Transformers + MLP \\
\cite{zhao2024lidar} & 2024 & Pose estimation & Transformers + MLP \\
\cite{kulkarni2025lidar} & 2025 & Object detection & PointNet \\
\cite{chappa2025ligar} & 2025 & Action recognition & PointNet + Transformers + MLP \\
\cite{meng2025indoor} & 2025 & Action recognition & CNN + LSTM \\

\hline
\end{tabular}
\begin{tablenotes}
\scriptsize
RPN-Region Proposal Network; AVOD-Aggregate View Object Detection; TD–Temporal Difference; GVF–General Value Function; NIFPN-Node-Iteration Fuzzy Petri Net.
\end{tablenotes}
\end{center}
    \vspace{-1cm}

\end{table}

\textcolor{black}{
\textbf{$\bullet$ Pose Estimation \& Deep Learning drive LiDAR-AI research:}
To better understand recent research directions in LiDAR-based AI solutions, a statistical synthesis of various studies have been compiled in Table~\ref{tab_ai_summary}. The findings show that pose estimation is by far the most frequently explored application, accounting for more than half of the reviewed studies. This is followed by gait recognition (around 22\%), which naturally aligns with LiDAR's ability to capture detailed movement. Other applications include action recognition (13\%), walking mode classification, arm segmentation, and full-body segmentation, each appearing in only a few works. The focus on body movement and posture tasks suggests that many rehabilitation-relevant areas are still underexplored. In terms of AI methods, deep learning dominates (87\%), with only 13\% of studies relying on traditional machine learning techniques. Among deep learning approaches, Transformer-based models with MLPs are the most common, followed by variants of PointNet, CNNs, and \textcolor{black}{recurrent neural networks (RNNs)} style encoders. This variety reflects active experimentation and growing interest across disciplines. It is worth noting that the number of studies in this field has been growing steadily in recent years, showing a clear increase in research interest. Still, remains relatively low, particularly in terms of practical applications in rehabilitation, underscoring the need for further research and real-world applications in the future.
}\\

\textbf{$\bullet$ LiDAR-only leads, multi-modal interest growing:} Fig.~\ref{graph1} illustrates the number of studies adopting multi-modal solutions over time. Although the Pattern exhibits year-to-year fluctuations, a general upward trend can be observed. However, their number still remains below half of the gathered the entire period, which means the reliance has been more on LiDAR-only solutions. LiDAR alone reduces the need for multiple sensors, lowering complexity, calibration requirements, and synchronization issues. On the other hand, fused modalities offer a broader view and have been shown in many works to outperform single-sensor solutions. Therefore, both approaches remain viable options for future research. 

\begin{figure}[t!]
    \centering
\includegraphics[width=0.9\linewidth]{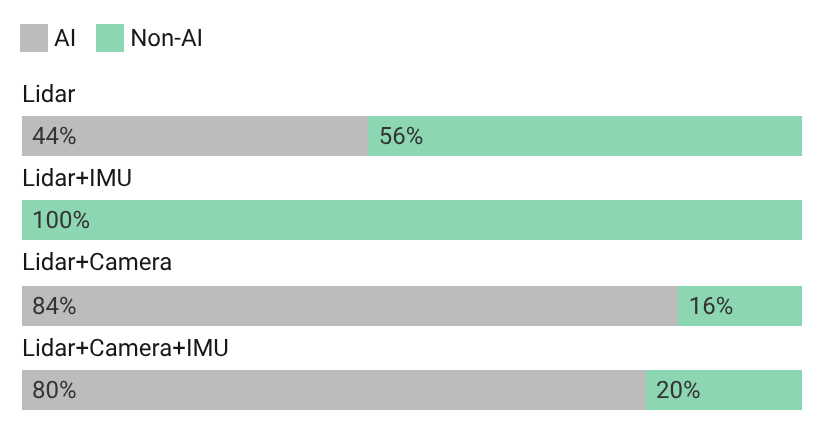}
    \caption{Percentage of studies using AI and non-AI approaches in relation to modality fusion: LiDAR alone, LiDAR with IMU, LiDAR with camera, and LiDAR with both camera and IMU.}
    \label{graph2}
    \vspace{-0.5cm}
\end{figure}

\textbf{$\bullet$ AI becomes essential for complex sensor fusion:} \textcolor{black}{Fig.~\ref{graph2} illustrates distribution of studies employing AI and non-AI methods across different modality fusion setups}, we notice that the majority of studies relying solely on LiDAR use non-AI solutions, mainly geometrical or mathematical algorithms. Still, around 44\% of the works adopt AI-based approaches, and we expect this to increase further in the future. On the other hand, when merging LiDAR with IMU data, no work was found to rely on intelligent algorithms or learning-based approaches, which represents a clear gap in the field. Conversely, when fusing point clouds with images, researchers often rely on AI approaches due to the growing complexity of processing and aligning point clouds with image pixels. Furthermore, when merging all three sensors, 80\% of the works we found rely on AI-based methods.
  
\begin{figure}[t!]
    \centering
\includegraphics[width=1\linewidth]{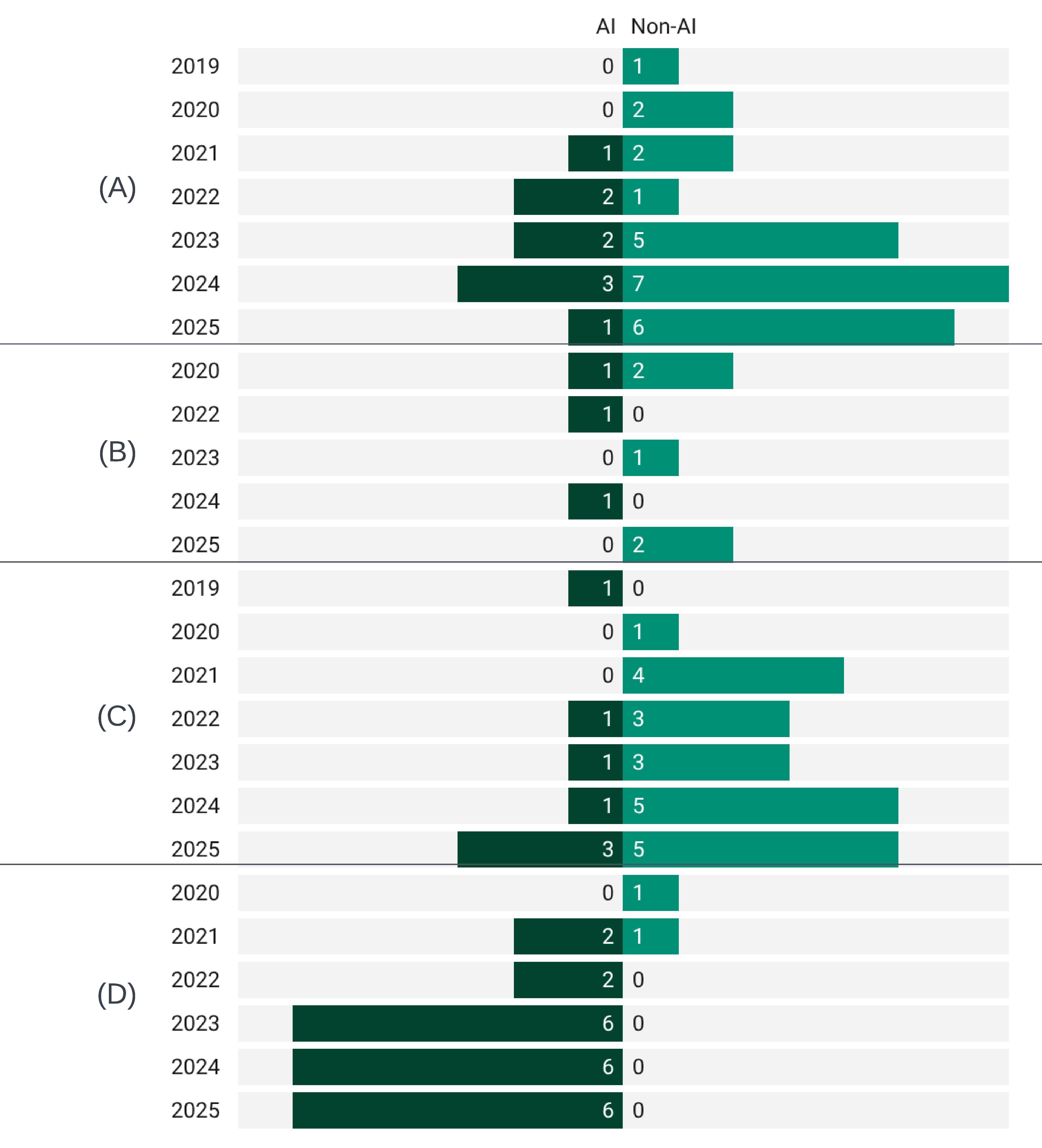}
    \caption{AI vs. non-AI approaches (2019–2024) in LiDAR-based healthcare applications: (A) Stationary LiDAR for Human Body Assessment, (B) LiDAR-enabled Rehabilitation Monitoring Robots, (C) LiDAR-enabled Navigation Assistive Robots, and (D) LiDAR-based Activity Recognition.}
    \label{graph3}
        \vspace{-0.7cm}

\end{figure}

\renewcommand{\arraystretch}{1.2} 
\begin{table*}[!h]
\renewcommand{\arraystretch}{1.2}
\centering
\caption{Summary of publicly available datasets in the literature that use LiDAR and are relevant for potential applications in healthcare and rehabilitation.}
\begin{tabular}{|@{\hskip3pt}p{2.2cm}@{}|@{\hskip3pt}p{3.2cm}@{}|@{\hskip3pt}p{0.7cm}@{}|@{\hskip3pt}p{3.5cm}@{}|@{\hskip3pt}p{1.4cm}@{}|@{\hskip3pt}p{1.4cm}@{}|@{\hskip3pt}p{4.8cm}@{}|}
\hline
\rowcolor{verylightgray}

\textbf{Category} & \textbf{Name} & \textbf{Year} & \textbf{Labels} & \textbf{\# Participants} & \textbf{\# Frames} & \textbf{Others} \\ 
\hline

\multirow{2}{*}{\textbf{LiDAR alone}} 
& PCG1, PCG2~\cite{yamada2020gait} & 2020  &  Participants ID & 30&-&-\\ \cline{2-7}
& Medical-Human-Pose~\cite{wu2024lidar} & 2024  &  3D pose annotations &-&-&3,752 depth images \\ 
\hline

& SUSTech1K~\cite{shen2023lidargait} & 2023  & Participants ID & 1,050 & 763,416& 25,239 sequences and 3,075,575 images\\ \cline{2-7}
& LiCamPose~\cite{cong2023weakly} & 2023  &   2D pose annotations & -& 47,470&-\\ \cline{2-7}
& LidPose~\cite{kovacs2024lidpose} & 2024  &  2D and 3D pose annotations &-&-& 9500 skeletons and 161,000 joints \\ \cline{2-7}
& Human-M3~\cite{fan2023human} & 2023 &  2D and 3D pose annotations & - &12,200& 89642 human pose\\ \cline{2-7}
\multirow[t]{9}{*}{\shortstack[l]{\textbf{LiDAR and}\\\textbf{camera}}}& FreeGait~\cite{han2024gait} & 2024  & Participants ID & 1,195&-&11,921 sequences\\ \cline{2-7}
& PeDX~\cite{kim2019pedx} & 2019  &  2D and 3D pose annotations & -& 2,500 &5,000 stereo images  \\ \cline{2-7}
& STCrowd~\cite{cong2022stcrowd} & 2022  &2D and 3D pose annotations & 219k & 10,891&- \\ \cline{2-7}
& Movin~\cite{jang2023movin} & 2023  &  3D pose annotations & 10 & 160K &-\\ \cline{2-7}
& HuMMan~\cite{cai2022humman} & 2022  &  3D pose annotations & 1,000& 60M& 400k sequences \\ 
\hline

& Sloper4d~\cite{dai2023sloper4d} & 2023  &  3D pose annotations & 12& 100k &300k video frames, 500k imu frames \\ \cline{2-7}
& MVPose3D~\cite{lee2025point2pose} & 2025  &  3D pose annotations &12 &215,039 & -\\ \cline{2-7}
& FreeMotion~\cite{ren2024livehps} & 2024  &  3D pose annotations &-& 78,775&-\\ \cline{2-7}
\multirow[t]{6}{*}{\shortstack[l]{\textbf{LiDAR, Camera,}\\\textbf{and IMUs}}}& LIDARHumman26M~\cite{li2022lidarcap} & 2022  &   3D pose annotations & 13 &184k &- \\ \cline{2-7}
& CIMI4D~\cite{yan2023cimi4d} & 2023 &  3D pose annotations & 12&180k&- \\ \cline{2-7}
& LIPD~\cite{ren2023lidar} & 2023   &  3D pose annotations & 15&808,608&- \\ 
\hline

\end{tabular} \vspace{0.5em} 
    \vspace{-0.5cm}
\label{datasets}
\end{table*}

\textbf{$\bullet$ AI excels in activity recognition, but rehabilitation robots remain underexplored:} Fig.~\ref{graph3} investigates the general research interest in different rehabilitation applications and compares the adoption of AI- versus non-AI-based solutions for each category. We specifically focus on:  
(A) stationary use of LiDAR for body scanning, either for limb modeling or gait monitoring;  
(B) LiDAR in rehabilitation robots, which track performed exercises and assess the correctness of movements; 
(C) LiDAR in assistive medical robots, helping impaired patients navigate smoothly; and  
(D) the use of LiDAR for activity recognition. The figure exposes a growing interest across these applications, except for category (B): For categories (A) and (C), we observe an increasing number of references over time, with more non-AI solutions being proposed.  
For category (B), there are very few works in the past six years, representing a gap in research. This application is particularly important in rehabilitation, as current movement tracking often relies on wearable sensors, tape-measurements, or caregiver observation, which are imprecise and prone to error. Therefore, it is crucial to develop robotic systems that closely follow exercise sessions and measure assessment parameters in real time with high precision.   
Finally, for category (D), the majority of proposed solutions are AI-based and have shown great success in position and activity recognition, as well as gesture identification.

\section{LiDAR Datasets Relevant to Rehabilitation}
We have carefully gathered the available datasets in the literature that use LiDAR. These datasets, although not specifically designed for rehabilitation, have been widely used to validate models in this field. The majority of them provide 3D point clouds, often accompanied by synchronized images and IMU data used as ground truth. The main annotations include either 2D or 3D key joints, positions, and even participants identification through gait analysis. In Table~\ref{datasets}, we summarize the relevant labeled datasets found in the literature. Important details are included, such as the type of labeling and information about the dataset size, including the number of persons involved and the number of point cloud frames it contains. With the growing use of AI, the need for publicly available data was frequently emphasized in many studies. We observe that the number of published datasets increased over the years, reaching a cumulative total of 12 by 2023. In 2024, however, a decline is noted, which may be explained by researchers now focusing on utilizing the already available datasets rather than introducing new ones.

Out of the 17 datasets presented in Table~\ref{datasets}, 12 are available online for download, which means a good number of datasets are publicly accessible. Most of these datasets are based on in-vivo data, collected in various real-world scenarios, such as Point2Pose (MVPose3D)~\cite{lee2025point2pose}, which focuses on real indoor environments. Only a few datasets include synthetic data: specifically, Medical-Human-Pose~\cite{wu2024lidar} and LIPD~\cite{ren2023lidar} datasets combine real and simulated point clouds. Most of the datasets are dedicated to position estimation or activity recognition, while only PCG1, PCG2~\cite{yamada2020gait}, and FreeGait~\cite{han2024gait} are suitable for subject identification. These datasets are relevant for rehabilitation applications because, beyond their original intended purposes, they include processed human body data, such as key joint annotations, extracted skeletons, or segmented body parts.

Some datasets are widely used as benchmarks for validating models. Notably, LIDARHuman26M~\cite{li2022lidarcap}, which includes 184k point cloud frames, is commonly used to evaluate position estimation models. This dataset was collected using a 3D LiDAR, a camera, and IMUs, and provides 3D keypoint annotations. Sloper4D~\cite{dai2023sloper4d}, collected from 12 participants, offers 100k annotated frames of 3D LiDAR data, 300k video frames, and 500k IMU data. Others include CIMI4D~\cite{yan2023cimi4d}, collected with an Ouster OS1 3D LiDAR, containing 180k annotated 3D frames, and the larger HuMMan~\cite{cai2022humman} dataset, which involves 1K participants and up to 60M synchronized point cloud and image frames. Finally, STCrowd~\cite{cong2022stcrowd} includes both 2D and 3D pose annotations, collected from 219k individuals, with up to 10,891 frames, containing multi-person skeleton annotations.

\vspace{-0.5cm}
\textcolor{black}{
\section{Challenges and Open Research Directions}}
\textcolor{black}{Although LiDAR has demonstrated strong potential in a wide range of rehabilitation applications, many challenges remain before it can reach its full capacity and be implemented effectively in practice. Current studies highlight limitations related to adaptability in clinical environments, integration with other systems, real-time performance, and the lack of suitable datasets. Addressing these challenges is essential for translating LiDAR from experimental use into practical rehabilitation solutions.}
\subsection{Enhancing LiDAR Reliability in Controlled Rehabilitation Environments} 
While LiDAR has demonstrated strong performance in various rehabilitation-related tasks, several challenges remain, indicating an open optimization problem for maximizing its potential. It is not feasible to simultaneously achieve all the advantages of this technology: longer ranges, wider fields of view, denser point clouds, multi-frame acquisition, lower computational complexity, high accuracy, real-time performance, full coverage using multiple devices, and low cost. These objectives involve inherent trade-offs.

For example, commercial devices can operate at long distances (up to 30–300 meters), but these specifications generally indicate only how far a few points can still be detected. At such distances, point density becomes very low. In typical rehabilitation applications, the practical effective range is at least three times shorter to ensure sufficiently detailed scans. Conversely, placing the LiDAR too close to the subject may lead to truncations in the point cloud. Sensor placement also affects accuracy: a tilted LiDAR can distort measurements due to its point-of-view perception. For instance, one study~\cite{single2024transferable} reported that when the LiDAR was positioned at chin level, the lower-limb dimensions appeared to be underestimated. Another example involves multi-LiDAR setups, which could provide multi-view scanning and a more complete 3D representation, as a single LiDAR cannot see behind objects. However, this approach requires careful temporal synchronization and spatial alignment of the data, and adding more sensors increases cost and complexity.

These trade-offs highlight a rich area for future research. Given that rehabilitation is performed in controlled environments, it is important to develop optimized strategies for LiDAR deployment in such applications.

\subsection{Overcoming LiDAR’s Occlusion Limits through Smarter Sensor Integration}
Multi-sensor solutions that include LiDAR are a highly promising approach for rehabilitation, where the main challenge lies in designing effective pipelines for data integration and fusion. LiDAR can be combined with other sensors, such as IMUs or different types of cameras. The multimodal data help overcome the limitations of single-sensor systems and strengthen decision-making by relying on multi-parametric inputs.

Data captured by each device must be precisely aligned in both spatial and temporal domains, requiring careful sensor placement, calibration during preprocessing, and proper temporal synchronization. Once aligned, the multimodal data can be fused and fed into processing models. Fusion can be performed at different levels. At the raw-data level, fusion can be done by concatenation or superposition, which increases input size and requires models capable of handling inputs of different types. Keeping each sensor’s data in its original format, such as point clouds for LiDAR or pixel matrices for images, makes raw-level fusion challenging. Therefore, it is important to represent the data in a uniform manner, for example, by mapping point clouds into pixel matrices or vice versa, or by bringing the coordinate system of IMUs into alignment with the LiDAR.

An alternative and often preferred approach is feature-level fusion: each sensor’s data is processed independently to extract features, which are then merged to produce a controllable and compact input for models that perform downstream tasks based on the combined feature set.
\subsection{Advancing Real-Time, Edge-Enabled LiDAR Systems in Rehabilitation}
LiDAR multi‑frame generated data is large in size and requires high computational power, large storage capacity, and complex processing. This complexity increases even further when using multiple LiDARs and/or combining LiDAR with other sensors, such as cameras, which adds more data streams that need to be processed simultaneously or successively. For this reason, it is strongly advised to focus on reducing data dimensionality and extracting the key features in the early stages of preprocessing in order to avoid an increased computational load throughout the long pipeline.

Furthermore, some applications require real‑time response, where delays may have serious consequences and can even be life‑threatening, whivh is particularly critical when the participants are in sensitive or critical conditions. Examples include detection of falling, navigation danger when assisting walkers, and sudden abnormal gait detection. For such applications, more focus must be directed toward the enabling of low‑latency response systems, which is challenging given the complexity and volume of the data that must be processed quickly. 

Most of the studies reported so far focus on trials conducted in the lab and do not extend their solutions to real‑life edge deployment scenarios, where constraints become much more critical. These include strict limitations on power consumption, response times, storage restrictions, and the robustness of the models in uncontrolled and dynamic real environments. For these reasons, real‑time and edge‑oriented deployment remains an important research direction in the context of LiDAR‑based rehabilitation applications.

\subsection{Developing Rehabilitation-specific LiDAR Datasets}
Besides the public availability, large volume, rich labeling, and data diversity of the existing datasets, they are still not dedicated to rehabilitation purposes. Rehabilitation applications have specific requirements, typically limited to controlled indoor environments and single-person scenarios. In contrast, most of the available datasets are designed for harsher conditions, multi-person interactions, and outdoor settings. This is because they are developed for other applications, such as human action recognition for autonomous driving. Such settings reduce the focus and precision on the human subject, and over the years the datasets proposed challenge the previous ones by adding more complexity, which makes them far from being suitable for rehabilitation applications.

Leveraging clean indoor spaces and focusing on a single person per frame allows achieving higher precision and denser point cloud data focused on the human body, which would make a dataset suitable for rehabilitation assessment. Such a dataset would contain point cloud frames as raw data, synchronized data from other sensors, and extracted key joint points. Labeling might include segmenting the data points, naming the key joints, and classifying the scans into different classes useful for supervised learning. Not all the samples must be labelled, the more the better, but with the advancement of AI techniques, semi‑supervised learning may leverage the use of both labelled and unlabelled data. This kind of dedicated dataset is still missing in the current literature.

\subsection{Enabling LiDAR-based Telerehabilitation and Home Monitoring}
LiDAR shows high accuracy in providing feedback for caregivers on the accuracy of therapy movements, joint angle measurements, and gait analysis. It also helps detect gait abnormality, dizziness, falls, long periods of movement stop, and gesture identification, for example to declare an emergency. All these applications enable telerehabilitation or home surveillance. It helps reduce the effort for patients to travel to rehabilitation centers and avoids moving patients for long periods to care facilities.

Using cameras at home discloses privacy, and using wearables is not comfortable and prone to errors. LiDAR in this case is the best solution to overcome all these limitations. With all the success that LiDAR‑based solutions show, more efforts must be directed toward applying these solutions at home to further simplify the rehabilitation process and make it easily accessible for people in need. Examples include video conferencing with real-time feedback from LiDAR-based systems, allowing remote contact between caregivers and patients. In addition, LiDAR-driven alarm and hazard-detection systems could support nighttime monitoring of elderly individuals or infants at home, ensuring rapid response and timely assistance when risks are detected.

\subsection{Exploring Generative AI for LiDAR Point Cloud Processing} 
Even with all the advantages offered by learning‑based models on LiDAR data and the high performance demonstrated in many studies, most existing processing techniques still rely on non‑learning approaches, mainly predefined algorithmic, mathematical, and geometry‑based methods. More focus needs to be directed toward proposing AI‑based approaches dedicated to processing point clouds. 
In addition, most of the current AI‑based solutions use models that were not originally designed for sparse point cloud data but rather for structured 2D data. Developing models specifically designed to take raw point clouds as input will reduce the burden of heavy preprocessing required to adapt the data to models that are not intended to handle such input. Competitive alternatives to PointNet and its improved version PointNet++ that directly process raw point clouds need to be designed to offer a broader range of solutions for this type of data.
Furthermore, no existing work has yet explored the potential of generative AI models in this context. These models have already shown great performance and robustness in various fields, especially in handling large volumes of text, images, tables, and documents, which share similarities with the large data produced by LiDAR. Exploring such possibilities remains an open and promising direction.

\section{Conclusion}
In this survey, we investigated recent trends in the use of LiDAR for rehabilitation applications, post-injury care, and hospital environments. We reviewed various applications, including the use of LiDAR as a standalone device or mounted on robotic systems for tasks such as body scanning, gait analysis, assistive devices, and pose and activity recognition. We presented the processing techniques employed in different studies, highlighting several notable trends. Additionally, we focused on advancements in AI techniques applied in this field, summarized the available datasets, and identified existing gaps.

We conclude that applications such as assistive robots, gait analysis, position estimation, and body scanning have received the most attention, whereas areas like real‑time rehabilitation monitoring and rehabilitation‑dedicated robots still require further exploration due to their significant potential. Current processing techniques are largely limited to geometrical, statistical, and mathematical approaches, as well as models originally designed for 1D or 2D data, highlighting the need for more solutions specifically dedicated to LiDAR data.

There is a clear need for datasets designed specifically for rehabilitation, focusing on the human body frame, low‑complexity environments, single-person scenarios, and normal lighting, with adequate labeling for supervised learning. For future work, more advanced fusion approaches at both the raw data and feature levels are needed as multimodal techniques gain attention, improving response time and addressing challenges related to data complexity and storage requirements. In addition, investigating generative AI for processing large-scale LiDAR data, as well as exploring the adaptability of LiDAR systems in rehabilitation environments and home settings, represents important directions for further research.

\bibliography{References}
\bibliographystyle{ieeetr}
\balance
\end{document}